\documentclass[authoryear, preprint,11pt]{elsarticle}

\usepackage{graphicx} 
\usepackage{booktabs} 
\usepackage{multirow} 
\usepackage{makecell} 

\usepackage{amssymb}
\usepackage{amsmath}
\usepackage{ragged2e}
\usepackage{color}
\usepackage{graphicx}
\usepackage{subcaption}
\usepackage{hyperref}
\usepackage{natbib}
\usepackage[dvipsnames]{xcolor}

\usepackage{geometry}
\usepackage[utf8]{inputenc}
\usepackage[english]{babel}
\usepackage{amssymb}
\usepackage{graphicx}
\usepackage{url}

\usepackage{algorithm}
\usepackage{algpseudocode}
\algrenewcommand\textproc{\text}
\usepackage{booktabs}
\usepackage{multirow}
\usepackage{enumitem}
\usepackage{balance}
\usepackage{makecell}
\usepackage{threeparttable}
\usepackage{amsmath,amsfonts,mathtools}
\usepackage{mathrsfs}
\usepackage{float}
\usepackage{enumitem}
\usepackage{adjustbox}
\usepackage{tabularx}
\usepackage{ragged2e} 
\usepackage{booktabs}
\usepackage{xcolor}
\usepackage{tabularx}
\usepackage{array}
\usepackage{amsmath}

\newcolumntype{Y}{>{\centering\arraybackslash}p{3cm}}

\begin{document}

\begin{frontmatter}

\author[inst1]{Xiaowei Gao}
\ead{xiaowei.gao.20@ucl.ac.uk}
\author[inst2]{Xinke Jiang}
\ead{XinkeJiang@stu.pku.edu.cn}
\author[inst3]{Dingyi Zhuang}
\ead{dingyi@mit.edu}
\author[inst4]{Huanfa Chen}
\ead{huanfa.chen@ucl.ac.uk}
\author[inst5]{Shenhao Wang}
\ead{shenhaowang@ufl.edu}
\author[inst6]{Stephen Law}
\ead{stephen.law@ucl.ac.uk}
\author[inst1]{James Haworth\corref{cor}}
\ead{j.haworth@ucl.ac.uk}

\cortext[cor]{Corresponding author}
\affiliation[inst1]{organization={SpaceTimeLab, University College London (UCL)},
            state={London},
            country={UK}}

\affiliation[inst2]{organization={School of Computer Science, Peking University (PKU)},
            state={Beijing},
            country={China}}
            
\affiliation[inst3]{organization={Department of Civil and Environmental Engineering, Massachusetts Institute of Technology (MIT)},
            state={Cambridge, MA},
            country={USA}}
\affiliation[inst4]{organization={The Bartlett Centre for Advanced Spatial Analysis, University College London (UCL) },
            state={London},
            country={UK}}
\affiliation[inst5]{organization={Department of Urban and Regional Planning, University of Florida},
            state={Gainesville, FL},
            country={USA}}
\affiliation[inst6]{organization={Department of Geography, University College London (UCL) },
            state={London},
            country={UK}}

\title{Uncertainty-Aware Probabilistic Graph Neural Networks for Road-Level Traffic Crash Prediction}

\newpage

\begin{abstract}

Traffic crashes present substantial challenges to human safety and socio-economic development in urban areas. Developing a reliable and responsible traffic crash prediction model is crucial to addressing growing public safety concerns and enhancing the safety of urban mobility systems. Traditional methods face limitations at fine spatiotemporal scales due to the sporadic nature of high-risk crashes and the predominance of non-crash characteristics. Furthermore, while most current models show promising occurrence prediction, they overlook the uncertainties arising from the inherent nature of crashes, and then fail to adequately map the hierarchical ranking of crash risk values for more precise insights. To address these issues, we introduce the \textbf{\underline{S}}patio\textbf{\underline{t}}emporal \textbf{\underline{Z}}ero-\textbf{\underline{I}}nflated \textbf{\underline{T}}wee\textbf{\underline{d}}ie \textbf{\underline{G}}raph \textbf{\underline{N}}eural \textbf{\underline{N}}etworks (STZITD-GNN), the first uncertainty-aware probabilistic graph deep learning model in road-level traffic crash prediction for multi-steps. This model integrates the interpretability of the statistical Tweedie family model and the expressive power of graph neural networks. Its decoder innovatively employs a compound Tweedie model—a Poisson distribution to model the frequency of crash occurrences and a Gamma distribution to assess injury severity, supplemented by a zero-inflated component to effectively identify exessive non-incident instances. Empirical tests using real-world traffic data from London, UK, demonstrate that the STZITD-GNN surpasses other baseline models across multiple benchmarks and metrics, including crash risk value prediction, uncertainty minimisation, non-crash road identification, and crash occurrence accuracy. Our study demonstrates that STZTID-GNN can effectively inform targeted road monitoring, thereby improving urban road safety strategies.

\end{abstract}


\end{frontmatter}

\newpage

\setlength{\parskip}{0.4em} 

\section{Introduction}
\label{sec:intro}
Traffic crashes persist as a major barrier to sustainable urban development \citep{dai2018evolving,tu2020portraying, barahimi2021multi,rodrigue20208}. The recent World Health Organisation (WHO) report reveals that road traffic crashes claimed approximately 1.19 million lives worldwide in 2021, making them the leading cause of mortality and disability and imposing an economic burden of up to US\$1.8 trillion, approximately 10–12\% of the global Gross Domestic Product (GDP)\citep{world2023global}. 

To support reductions in traffic crashes, there is a growing emphasis on the development of models aimed at predicting the location, severity, and causes of crashes. The early approaches relied mainly on linear regression models \citep{mountain1996accident,greibe2003accident}, models that incorporate random parameters \citep{el2009accident,dinu2011random}, and geographically weighted regression (GWR) models \citep{li2013using,xu2015modeling} to predict regional traffic crashes. Recent research has turned to deep learning-based methods, notably graph neural networks (GNNs), which are able to handle large datasets and capture complex spatial and temporal representations of urban road networks. Examples include time-varying GNN \citep{zhou2020riskoracle}, multiview fusion GNN\citep{trirat2023multi} and subgraph-based GNN\citep{zhao2023gmat}. Despite substantial advancements in model development, accurately predicting road crashes at a fine spatiotemporal granularity road presents several challenges:

\begin{itemize}
    \item \textbf{Spatially Imbalanced Crash Occurrence:}The prediction of road crashes is challenging due to the uneven spatial distribution of crashes. Firstly, zero-inflation is inherent to crash data, where areas with no crashes are much more common than those with crashes. This spatially skewed distribution causes models to be biased toward predicting zero crashes, complicating the effective training of predictive algorithms\citep{wu2022multi}. Secondly, where crashes do occur, they are predominantly minor, leading to an abundance of low-risk scores across the spatial grid. This creates a misleading representation that underestimates the potential severity of less frequent, more severe crashes\citep{shirazi2019characteristics, saha2020application}. The combination of these factors—widespread zero-crash areas and the dominance of minor crashes in crash-prone zones—results in a challenging scenario for models that need to understand and predict the full spectrum of crash severities across different areas \citep{wang2023uncertainty}.
    \item \textbf{Temporal Constraints with Uncertainty:} Building on the spatial challenges, the temporal dynamics of traffic crashes introduce additional complexity. Particularly, predictive models for traffic crashes face the challenge of accurately forecasting not just immediate but also future conditions, necessitating reliable multi-step predictions \citep{soltani2024space}. The nature of traffic crashes, characterized by their non-linearity and the variable time intervals between events, adds complexity to this task. As the prediction horizon extends, uncertainty increases, particularly because future traffic conditions and potential crash triggers can change substantially over time\citep{qian2022uncertainty}. Most current deep learning methods do not fully address these challenges. They tend to overlook the complexities of aleatoric uncertainty, which arises from the inherent randomness in the data. These models often rely on assumptions of uniform variance across all data points, producing point estimates from average outcomes\citep{wang2023uncertainty, zhuang2022uncertainty, jiang2023uncertainty}. This approach fails to capture the temporal variability and the unpredictable nature of crashes, resulting in predictions that do not adequately reflect the true range of future possibilities. 
  
\end{itemize}

To address the intricate challenges posed by the spatial and temporal distribution of crash data, we introduce the Spatio-temporal Zero-Inflated Tweedie Graph Neural Network (STZITD-GNN). This model combines spatio-temporal graph machine learning with statistical methods, enabling enhanced uncertainty quantification over multi-step prediction intervals. It is specifically designed to handle the probabilistic assessment of non-recurrent, road-level traffic crashes, providing a robust solution to predict spatially and temporally varying crash data.

The proposed framework integrates a cohesive encoder that combines a Gated Recurrent Unit (GRU) for capturing temporal dynamics and a Graph Attention Network (GAT) for spatial relationships, ensuring a synchronised capture of spatio-temporal dependencies across individual roads and time steps. The decoder leverages a Tweedie (TD)-based distribution, a flexible compound Poisson-gamma model that simultaneously models both exact zero occurrences and continuous positive values. The choice of the TD distribution is motivated by its established efficacy in fields challenged by nonnormal distributions, such as crash frequency estimation \citep{wang2019functional}, insurance claims \citep{smyth2002fitting}, actuarial studies \citep{shi2016insurance}, and meteorological precipitation \citep{dunn2004occurrence}. This approach allows the model to output the TD distribution parameters, explicitly determining the upper and lower bounds for the prediction intervals of each sample, thus providing a probabilistic representation of each road’s crash risk scores.

Given the inherent extreme imbalance that traditional TD models struggle to adequately capture \citep{khosravi2010lower, abe2017non}, we have enhanced our model by incorporating an additional parameter, $\pi$, to estimate the probability mass of zero values. This adjustment leads to the creation of the Zero-Inflated Tweedie (ZITD) distribution on the decoder, which offers a more comprehensive and responsible depiction of real crash scenarios. This refinement ensures that our model not only captures the complexities of traffic crash data, but also enhances the accuracy and reliability of its predictions. Specifically, our main contributions can be summarised as follows.

\begin{itemize}
	\item To address the limitations of ST-GNN in modelling complex spatial and temporal relationships, our framework introduces an end-to-end spatio-temporal encoder. This encoder is specifically designed to simultaneously model pairwise relationships and integrate them into a cohesive embedding. This approach enhances the ability of our model to capture and synthesize the intricate dynamics between spatial and temporal data points. 

	\item An innovative ZITD distribution is employed within our model to effectively address the uncertainties associated with zero-inflation and the long-tailed characteristics prevalent in road crash data. This approach enhances the model's capability to accurately represent the complex distribution patterns observed in traffic crash data.

	\item To the best of our knowledge, the STZITD-GNN is the first work to use probabilistic framework for granular, multistep road-level crash forecasting. The use of Likelihood-based Loss minimization rather than the tradional regression ones enhances its uncertainty-aware predictive accuracy and reliability.

	\item Extensive testing with real-world data confirms the superior performance of our framework compared to existing state-of-the-art approaches. Notably, it achieves a reduction in regression error of up to 34.60\% in point estimation metrics and shows a significant improvement, increasing at least by 55.07\% in uncertainty-based metrics.

\end{itemize}

This manuscript is structured as follows. In Section \ref{sec:lr}, we review spatiotemporal modelling methodologies in predicting traffic risk. Section \ref{sec:method} provides a detailed description of the proposed model. Section \ref{sec:experiements} compares our empirical results with state-of-the-art benchmark models, further enhancing our assertions with illustrative visualisations. In Section \ref{sec:conclusions}, we conclude the article by highlighting the advantages of our model and suggesting avenues for future improvements.

\section{Literature Review}
\label{sec:lr}
Predicting traffic crashes is essential for urban planning and mobility management. Initially, research in this area predominantly utilised traditional statistical methods and classical machine learning techniques, such as regression models \citep{chang2005analysis,lord2005poisson,caliendo2007crash,bergel2013explaining}, Bayesian networks \citep{martin2009bayesian,mujalli2011method,hossain2012bayesian}, tree-based models \citep{chong2004traffic,chang2005data,wang2010incident,lin2015novel}, and the k-nearest-neighbour method \citep{sayed1998comparison,tang2005traffic,lv2009real}. These early studies were generally confined to small geographical areas and demonstrated limited effectiveness in capturing the nonlinear and dynamic spatio-temporal patterns of the contributing factors \citep{chang2005data, anderson2009kernel,zhang2014crash}. In addition, these methods often analyse crash data in isolation, neglecting critical interdependencies between different locations \citep{wang2013effect,wang2021gsnet}. As a result, their broader applicability to city-wide analysis with large data size was constrained, highlighting the need for more advanced models capable of addressing these limitations.

Recent advances in deep learning techniques have shown promise in urban traffic crash prediction by incorporating spatio-temporal characteristics of the data. Convolutional neural networks (CNNs), recurrent neural networks (RNNs), and graph-based networks, such as graph-convolution networks (GCNs), have demonstrated their effectiveness in capturing intricate spatial patterns, capturing temporal dependencies, and modelling complex interactions \citep{wang2021ai,liu2020urban}. As summarised in Table \ref{tab:lr_summary}, \citet{chen2016learning} were among the first to explore city level crash prediction employing a stack denoise autoencoder. They integrated human mobility GPS data and historical incident point data at the grid level to map the real-time crash situation in Tokyo. However, while their approach attempted to capture a larger spatial area, it did not consider urban geo-semantic information for precise and long-term crash prediction. Later, \citet{chen2018sdcae} introduced a Stack Denoise Convolutional Autoencoder that incorporated spatial dependencies using stacked CNNs. Nonetheless, both of these studies overlooked the influence of temporal factors. Although RNNs are capable of handling temporal information, they are more suitable for short-term temporal learning \citep{ren2018deep}. To better address the temporal aspect, \citet{ren2018deep} used Long-Short-Term Memory (LSTM) to consider influential temporal factors between several locations. Their work represented the first attempt to incorporate temporal factors into deep learning models for the prediction of city-wide traffic crashes. Similarly, \citet{moosavi2019incident} employed LSTM networks for country-wide crash prediction, demonstrating their efficacy in predicting a binary outcome. Building on the LSTM framework, \citet{yuan2018hetero} proposed a more advanced model known as the Convolutional Long Short-Term Memory (Hetero-ConvLSTM) neural network based on spatial heterogeneous data. This model aimed to jointly address spatial heterogeneity and capture temporal features in state-wide crash prediction. However, their approach relied on pre-defined moving windows based on the spatial features of Iowa, United States, limiting its generalisability to other cities. \citet{bao2019spatiotemporal} integrated CNNs, LSTM as well as ConvLSTM into a spatiotemporal convolutional long short-term memory network (STCL-Net). Their work presented a more practical and feasible approach for the prediction of city-wide traffic crashes. By combining the strengths of these models, STCL-Net effectively captured spatiotemporal dependencies in crash data and influential factors, leading to improved prediction performance. 

To capture local and global dynamics with hierarchical spatial information, \citet{zhu2019ta} proposed the Deep Spatio-temporal Attention Learning Framework, which incorporates a spatio-temporal attention mechanism to effectively capture the dynamic impact of traffic crashes at different spatial levels. One notable aspect of their work was the emphasis on using real traffic administrative areas instead of manually divided grids. However, this approach does not incorporate spatial relationships, which limits its capacity to model the spatial heterogeneities that characterise traffic crashes. In a similar vein, \citet{huang2019deep} introduced the Deep Dynamic Fusion Network (DFN) framework, which aggregates heterogeneous external factors across both spatial and temporal dimensions. Using a temporal aggregation layer, the DFN framework is designed to automatically capture external influences from latent temporal dimensions.

Although the mentioned studies effectively captured temporal and spatial patterns in traffic crash data, yielding promising prediction outcomes, their lack of consideration of latent spatial correlations within urban structures can disrupt regional patterns in practical urban science and lead to prediction inaccuracies. Therefore, Graph-based networks have become a key method for city-wide traffic crash prediction by leveraging spatial graph features from actual urban networks \citep{xue2022quantifying, cheng2022network, li2022mining, SPGCL}.

\citet{zhou2020riskoracle} developed the RiskOracle framework by constructing an urban graph of regions and integrating the Differential Time-varying Graph Neural Network (DTGN) to identify dynamic correlations between historical traffic conditions and risks, enabling simultaneous predictions of traffic flow and crashes. To tackle zero-inflation in crash data, they introduced a data enhancement strategy based on a priori knowledge (PKDE), defining subregional crash labels for training. This approach adjusts zero crash values to a range between 0 and 1, maintaining the hierarchy of actual crash levels. Expanding on this foundation, \citet{zhou2022foresee} refined the DTGN framework by embedding an LSTM-based hierarchical sequence learning architecture, accommodating multiscale historical spatio-temporal data for nuanced long-term and short-term predictions. 
\citet{wang2021gsnet} and \citet{wang2021traffic} both focus on the integration of geographical and semantic relationships through different graph structures. While \citet{wang2021gsnet} used crash, road, and POI graphs to capture regional semantics and introduced modules for geographical and semantic spatio-temporal correlations, \citet{wang2021traffic}) combined attention-based LSTM, CNN, and GCN for multi-view predictions of traffic crashes at various scales. Although both approaches tackled the zero-inflation issue with promising results, they share limitations in addressing dynamic geographical semantics and alignment issues in graph structures. In contrast, \citet{trirat2023multi} proposed a multi-view graph neural network incorporating dynamic and static similarity information, offering a more dynamic approach to understanding traffic crashes. Using the Huber loss function, a parameter-based regression loss, their model aimed at robust regression that effectively accommodates zero inflation, showcasing an advanced strategy for precise prediction of urban traffic crashes.

Recent attempts to apply graph-based deep learning for road-level crash prediction have shown promise, but also highlight areas for improvement. \citet{yu2021deep} developed a Deep Spatiotemporal Graph Convolutional Network with a unique three-layer network, treating the road graph, spatio-temporal data, and embeddings separately. They tackled zero inflation through undersampling, balancing the dataset between risky and non-risky roads, and framing crash prediction as a binary classification. While this method effectively predicts crashes at the road level, the binary outcome overlooks the varying degrees of crash risk across roads. Similarly, \citet{wu2022multi} designed a fully connected dynamic network to predict road-level crashes by focussing on dynamic spatio-temporal correlations. To mitigate zero inflation, they introduced a cost-sensitive learning module that enhances positive sample classification accuracy by considering the imbalance between zero and non-zero crash values. However, this work only considered limited road segments in New York City with binary crash occurrence outputs, highlighting the need for further development to capture the full spectrum of crash severity and provide a more detailed understanding of road-level crash scenarios.

In summary, despite significant progress in urban traffic crash prediction yielding promising results, existing studies have yet to fully address uncertainty quantification in their predictions and provide multi-step crash assessments at the detailed road level.

\begin{table}[htbp]
  \centering
  \vspace{0pt}
  \caption{Summary of Deep Learning Methods in Traffic Crash Prediction}
  \footnotesize
  \renewcommand{\arraystretch}{1.5} 
  \begin{tabular}{>{\raggedright\arraybackslash}p{3cm}@{\hspace{5pt}}p{5.5cm}@{\hspace{5pt}}p{2cm}@{\hspace{2pt}}p{1.5cm}@{\hspace{2pt}}p{2cm}@{\hspace{2pt}}p{2cm}}
    \toprule
    \textbf{Reference} &  \textbf{Methodology} & \textbf{Spatial Unit} & \textbf{Zero-Inflation Issue} & \textbf{Prediction Steps} & \textbf{Prediction Outcome} \\
    \midrule
    \citet{chen2016learning} & Stack Denoise Autoencoder &  Regions & N/A & Single & Crash Risk Score \\    
    \citet{ren2018deep} & Long Short-term Memory network & Regions & N/A & Multiple & Crash Risk Score\\       
    \citet{chen2018sdcae} & Stack Denoise Convolutional Autoencoder &  Regions & N/A & Single & Crash Risk Score \\    
    \citet{yuan2018hetero} & Convolutional Long Short-Term Memory & Regions & N/A & Multiple & Crash Risk Score\\    
    \citet{moosavi2019incident} & Long Short-term Memory Network &  Regions & N/A & Single & Crash Occurrence \\    
    \citet{zhu2019ta} & Deep Spatial-Temporal Attention Learning Framework & Regions & N/A & Multiple & Crash Risk Score  \\
    \citet{huang2019deep} & Dynamic Fusion Network &   Regions & N/A & Single & Crash Occurrence\\    
    \citet{bao2019spatiotemporal} & Spatiotemporal Convolutional LSTM & Regions & N/A & Single & Crash Risk Score \\    
    \citet{zhou2020riskoracle} & DTGN with PKDE &  Regions & Yes & Single & Crash Risk Score \\      
    \citet{wang2021gsnet} & GSNet with Weighted Loss & Regions & Yes & Single & Crash Risk Score \\    
    \citet{wang2021traffic} & MVMT-ST Networks & Regions & Yes & Single & Crash Risk Score \\
    \citet{yu2021deep} & Deep ST-GCN with Undersampling & Roads & Yes & Single & Crash Occurrence \\
    \citet{wu2022multi} & MADGCN with Cost-sensitive Loss & Roads & Yes & Multiple & Crash Occurrence\\
    \citet{zhou2022foresee} & Enhanced DTGN & Regions & Yes & Multiple &  Crash Risk Score\\    
    \citet{trirat2023multi} & MV-GNNs with Huber Loss & Regions & Yes & Multiple & Crash Risk Score \\
    \bottomrule
  \end{tabular}%
  \label{tab:lr_summary}%
\end{table}%

\section{Methodology}
\label{sec:method}
In this section, we first introduce the concepts and define the framework for predicting road-level traffic crashes in Section \ref{Preliminaries and problem definition}. We then illustrate the statistical innovations from the TD distribution to ZITD in the context of uncertainty quantification in Section \ref{sec:from}. Subsequently, we introduce our new STZITD-GNN model, a spatio-temporal probabilistic modelling framework that combines the strengths of GRU and GAT in Section \ref{Framework of STGNNs}. The section ends with an in-depth exploration of the loss function employed within our model in Section \ref{3.4}. The major notations used in this section are listed in Table~\ref{tab:symbols}. 

\begin{table}[!htbp]
 \centering
 \footnotesize
 \caption{Main notations in this paper}
 \begin{tabular}{l|l}
  \toprule
  \textbf{Notation} &  {\textbf{Description}} \\
  \midrule
        $Y~/~X$             & Road-level traffic crash risk score ~/~ Road-level spatiotemporal traffic feature\\
        $V~/~E~/~A$       & Road set ~/~ Edge set ~/~ Road adjacency matrix\\
        $v_{i} ~/~ (v_{i}, v_{j})$ & The $i^{th}$ road ~/~ Edge between the $i^{th}$ road and $j^{th}$. \\ 
        $N ~/~ T$               & Numbers of roads ~/~ Time length\\
        $C_{it}$               & Number of crashes on road $v_i$ at time slot $t$\\
        $l_{it}^{j}$               & The $j^{th}$ cardinality values of crash severities in the time slot $t$ for road $v_i$\\
        $d ~/~ F ~/~ F'$               & Feature dimension/Temporal feature dimension/Spatio-Temporal feature dimension\\
        $\mathcal{ST}$ & Spatiotemporal deep learning Model \\
        $f_{\text{TD}}(\cdot)$ & Probabilistic density function of Tweedie distribution \\
        $\rho /~ \phi /~ \mu /~ \pi$ & Index parameter/ Dispersion parameter/ Mean parameter/ Zero-inflated parameter \\
        $\theta ~/~ \lambda ~/~ \alpha, \gamma$ & Natural parameter ~/~ Poisson mean parameter ~/~ Gamma parameters \\
        $\mathcal{Z}_T ~/~ \mathcal{Z}$ & Temporal embedding ~/~ Spatiotemporal embedding\\
       $\text{ReLU}(\cdot) /~ \sigma(\cdot) /~ \text{LeakyReLU}(\cdot) /~  [\cdot||\cdot]$             & ReLU activation /  Sigmoid activation / LeakyReLU activation /  Concatenation \\
       $W_r, W_u, W_c, W_a, b_r, b_u, b_c, \vec{a}$ & Learnable weight matrics of GRU and GAT\\
       $W_{\pi}, W_{\mu}, W_{\phi}, W_{\rho}, b_{\pi}, b_{\mu}, b_{\phi}, b_{\rho}$ & Learnable weight matrics of the four Parameter Encoders\\
       $h_t ~/~ r_t ~/~ u_t$ & Hidden feature ~/~ Reset gate ~/~ Update gate \\
       $\alpha_{i,j}$ & Attention value between road $v_i$ and $v_j$ \\
       $M$  & Number of attention heads\\ 
       $\epsilon$ & The minimum value \\
       $\eta$ & L2 normalisation weight-parameter \\
  \bottomrule
 \end{tabular}
 \label{tab:symbols}
\end{table}

\subsection{Preliminaries and Problem Definition} \label{Preliminaries and problem definition}
This section presents the basic concepts and terminology relevant to the study, followed by a description of the problem.

\noindent \textbf{Definition 1. \textit{Road Traffic Crash Risk Values}}: A traffic crash refers to an crash that involves one or more road users that leads to physical injury, loss of life, or property damage. We denote the set of roads as $V$ of size $N=|V|$, where $v_{i}$ signifies the $i^{th}$ road. The traffic crash risk on the roads is represented as $Y\in \mathbb{R}^{N\times T}$ for a duration of total $T$ periods, and $y_{it}$ denotes the crash risk in the $t^{th}$ time slot for the road $v_i$. Thus, $Y_{t}\in \mathbb{N}^{N}$ signifies the crash risk for all roads in the $t^{th}$ time interval, where $y_{it}$ is its constituent element. The calculation of traffic crash risk score $y_{it}$ for the road $v_i$ at the time slot $t$ is given by:

\begin{equation}
\centering
 \begin{aligned}
& y_{it} = \sum_{j=1}^{3} C_{i,k}^{t} \times l_{j},
    \label{eq: y_it}
 \end{aligned}
\end{equation}
where $C_{i,k}^{t}$ denotes the total number of $k$ crashes with the severity level $l_{j}$ at time slot $t$ for road $v_i$. We allocate the crash point to its closest road. As per prior research \citep{wang2021gsnet,trirat2023multi}, $l$ is assigned the values 1, 2, and 3, representing minor injury, serious injury, and fatal crash severities, respectively. $y_{it}$ equals 0 if there are no traffic crashes.

\noindent \textbf{Definition 2. \textit{Spatio-temporal Features}}: The embedding features are represented as $X \in \mathbb{R}^{N \times T \times d}$, where $d$ is the dimension of the features. We denote $x_{it}$ as the spatio-temporal feature of the road $v_i$ at the time slot $t^{th}$. Consequently, $X_t\in \mathbb{R}^{N\times d}$ signifies the features of all roads at time slot $t$. Our study includes spatial features such as road types, road lengths and widths, road conditions, and census characteristics for each lower layer super output area (LSOA) of roads. Temporal features, which vary daily, encompass weather information such as sunrise and sunset times, humidity, visibility, rainfall, etc., along with the identification as a holiday or a working day.

\noindent \textbf{Definition 3. \textit{Road Connection Graph}}: The road network is represented as a graph $\mathcal{G} = (V, E, A)$. Here, $E$ denotes the set of edges and $A\in \mathbb{R}^{N\times N}$ is the adjacency matrix that describes the connections between roads. An entry $A_{i,j}=1$ indicates the existence of an edge between road $v_i$ and $v_j$, denoted as $(v_i, v_j)\in {E}$, while $A_{i,j}=0$ implies no connection, denoted as $(v_i, v_j) \notin {E}$. 

\noindent \textbf{Problem Definition}: Our model aims to predict the future crash risk score in the next $p$ time windows and the confidence interval of the predicted results per road. It is to use historical records $X_{1:t}, Y_{1:t}$ and the graph structure $A$ as input data for training, to predict the probabilistic density function $f(Y_{t+1:t+p})$ of the distribution of $Y_{t+1:t+p}$.

\begin{equation}
\mathcal{ST}_\Theta([X_{1:t}, Y_{1:t}], A) \to f(Y_{t+1:t+p}),
    \label{eq: prediction objective}
\end{equation}
where $\Theta$ denotes the learned parameters for the ST-GNN model (also referred to as $\mathcal{ST}$). The multistep predictions are made for a 14-day output, as $p=14$.

\begin{figure*}[t]
\vspace{-0.48cm}
    \centering  \includegraphics[width=0.99\linewidth]{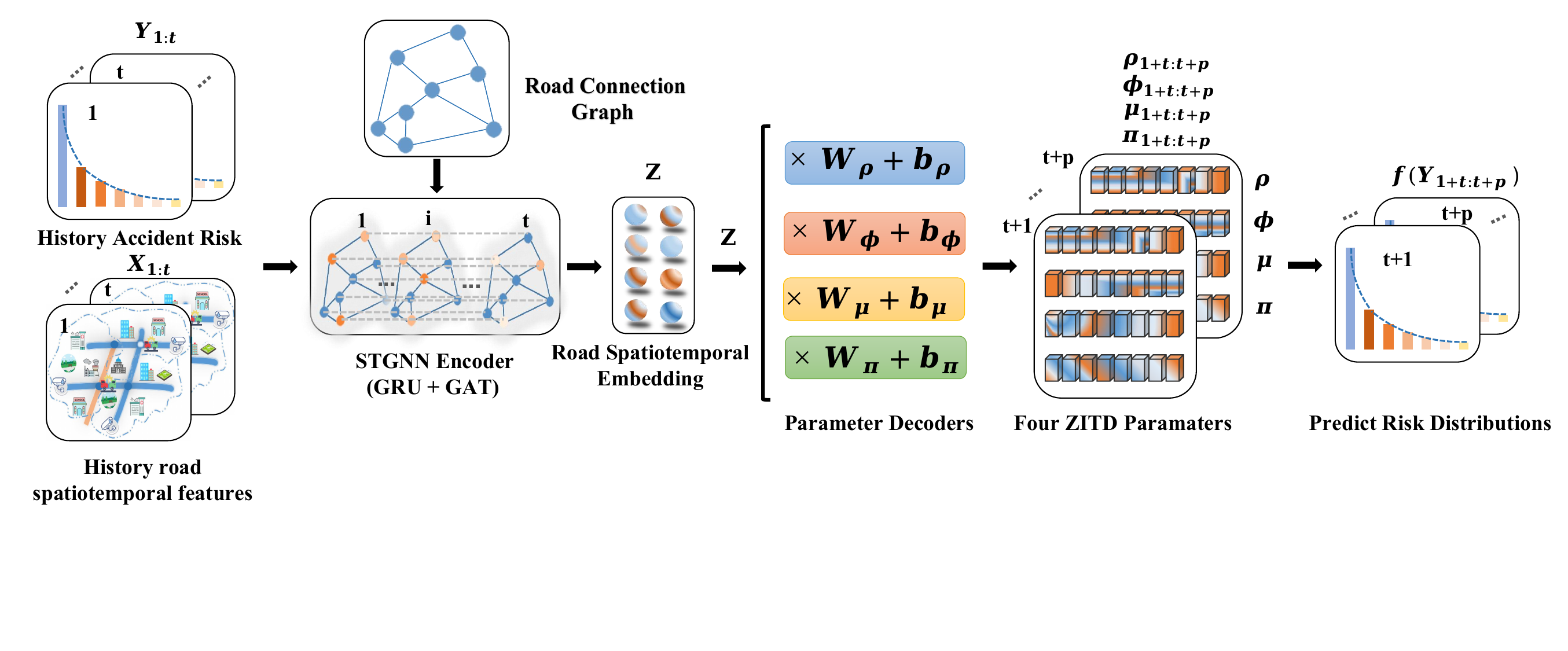}
    \vspace{-1.3cm}
\caption{The overall framework of STZITD-GNNs. STZITD-GNNs utilise the ST-GNN Encoder $\mathcal{ST}$ (composed of GRU and GAT encoders) to encode the history time window $1:t$ crash risk $Y_{1:t}$ and road features $X_{1:t}$ with the use of the road connection graph in the spatial-temporal embedding of the road $ \mathcal{Z}$. After encoding, the four parameter decoders map $\mathcal{Z}$ into the ZITD parameter space and obtain $\pi_{t+1:t+p}, \mu_{t+1:t+p}, \phi_{t+1:t+p},\rho_{t+1:t+p}$ for the predicted time window $t+1:t+p$, which determine the predicted road traffic crash risk distribution $f(Y_{t+1:t+p})$.}
    \label{fig:method}
\end{figure*}

\subsection{From TD Distribution to ZITD Distribution}
\label{sec:from}
This subsection provides an in-depth exploration of the progression from the TD to the ZITD distribution in the specific context of road crash prediction. Building on the work of \citet{lord2010statistical} and \citet{smyth2002fitting}, we recognise the effectiveness of TD distribution to address overdispersed data, indicating its potential applicability in multi-step road-level traffic crash prediction.

As illustrated in Section \ref{Preliminaries and problem definition}, when we aggregate crash data (incorporating various severities) from different roads or identical roads over different time intervals, the resulting distribution of crash counts tends to exhibit overdispersion. In detail, the crash value applied to both crash counts and the associated severity - is the prevalence of 'excess' zeros, implying more zero entries than would be predicted under a corresponding Poisson process. Meanwhile, in nonzero crash values, a clearly left-skewed but long-tailed distribution presents a smaller proportion of higher severity crashes. In these scenarios, the variance of the crash counts outweighs the mean crash count, creating a deviation from the Gaussian distribution, as underlined by the work of \citet{halder2019spatial} and \citet{wang2023uncertainty}. 

\subsubsection{TD Distribution} \label{sec:Tweedie (TD) Distribution}
In accordance with Equation \ref{eq: y_it}, we designate $y_k$ as the crash risk value of the road $i$ on time slot $t$. As a special case of Exponential Dispersion Models (EDMs), the TD model encompasses a broad family of statistical distributions \citep{tweedie1984index,jorgensen1987exponential,jiang2023uncertainty}. A random variable that follows the TD distribution has a probability density function $f_{TD}$ \citep{jorgensen1987exponential}, as follows :

\begin{equation}
f_{\text{TD}}(y_k|\theta, \phi) \equiv a(y_k, \phi) \exp {\bigl[ \frac{y_k(\theta)-\kappa(\theta)}{\phi} \bigl]},
\label{eq: td distribution}
\end{equation}
here, $\theta \in \mathbb{R}$ represents the natural parameter, while $\phi \in \mathbb{R}^+$ stands for the dispersion parameter. The normalizing functions $a(\cdot)$ and $\kappa(\cdot)$ are parameter functions for $\phi$ and $\theta$, respectively, detailed further in the subsequent discussion. For a TD distribution of EDMs, the mean and variance of variable $y_k$ are given by:

\begin{equation}
E(y_k) = \mu = \kappa(\theta)^{'}, \quad Var(y_k) = \phi \kappa(\theta)^{''},
\label{eq: td mean, var}
\end{equation}

In these equations, $\kappa(\theta)^{'}$ and $\kappa(\theta)^{''}$ represent the first and second derivatives of $\kappa(\theta)$, respectively. Meanwhile, the function $\mu\geq 0$ works as the mean parameter. Since the mean $E(y_k)$ and variance $Var(y_k)$ are dependent, $\kappa(\theta)^{''}$ can be directly expressed as $V(\mu)$, thereby providing a direct mean-variance relationship \citep{saha2020application,mallick2022differential}. Additionally, the TD family encompasses a number of vital distributions for distinct index parameters $\rho$ as demonstrated by $V(\mu)= \mu^ \rho $. These include the Normal ($\rho=0$), Poisson ($\rho=1$), Gamma ($\rho=2$), Inverse Gaussian ($\rho=3$), and Compound Poisson-Gamma distribution ($1<\rho<2$) and could be referred to as \ref{app:td table}. The Compound Poisson-Gamma distribution is particularly applicable, given its focus on parameterizing zero-inflated and long-tailed data, which aligns with the observed characteristics of real traffic crash data. The $\mu, \phi$ and $\rho$ are the three parameters that govern the probability and expected value of the risk of traffic crashes, which can be found from \ref{app:parameter}.

\subsubsection{ZITD Distribution}

The TD model, introduced in Section \ref{sec:Tweedie (TD) Distribution}, is designed to account for the occurrence of zeros by allowing a positive probability of zero outcomes, as demonstrated by formula $\mathbb{P}(y_{k}=0)=\mathbb{P}(C_{k}=0)=\exp (-\lambda)$. However, this model struggles to accurately represent crash data on roads where zero values overwhelmingly dominate, leading to an extremely unbalanced distribution that can misrepresent the actual crash rate \citep{mallick2022differential,zhou2022tweedie}.

To address this issue, we introduce a refinement, known as the Zero-Inflated Tweedie (ZITD) model with Four-parametric compound distributions. The ZITD model includes a distinctive feature, a sparsity parameter ($\pi$) that specifically addresses the skewness of the data characterised by an excess of zeros. The parameter $\pi$ signifies the likelihood of zero inflation, effectively distinguishing between zero and nonzero occurrences. In contrast, the complement probability ($1-\pi$) aligns with the traditional TD, facilitating a balanced representation of the crash risk values across the spectrum.
\begin{equation}
    \label{eq:zitd}
    y_k = \left\{ \begin{array}{cl}
        0 &  \text{with probability } \pi  \\
        Y &  \text{with probability } 1-\pi, Y\sim f_{TD}(Y|\mu, \phi, \rho)
    \end{array}\right..
\end{equation}

Finally, denoting the ZITD model as $y_k\sim f_{ZITD}(\pi, \mu, \phi, \rho)$, we formulate the probability density function of this model as follows:

\begin{equation}
    \label{eq:zitd_den}
    f_{ZITD}(y_k|\pi,\mu, \phi, \rho) = \left\{ \begin{array}{cc}
        \pi + (1-\pi)f_{TD}(y_k=0|\mu, \phi, \rho) & \text{if } y_k=0  \\
        (1-\pi)f_{TD}(y_k|\mu, \phi, \rho) & \text{if } y_k>0
    \end{array}\right..
\end{equation}

In accordance with this specification, we tackle the problem of extreme zero inflation by applying a linear weighting to the zero value, such that $\mathbb{P}(y_k=0)=\pi + (1-\pi) \exp{(-\frac{1}{\phi} \frac{\mu^{2-\rho}}{2-\rho})}$. This probability is significantly higher than the probability of observing a zero count from the TD distribution, intuitively indicating a substantial zero inflation. This modification ensures that the mean value of the ZITD distribution is $\mathbb{E}(y_k)= (1-\pi)\mu$. 

In the ZITD model, each parameter plays a distinct role in representing road-level crash risks. The sparsity parameter ($\pi$) is adjusted for instances without crashes, significantly influencing the dispersion parameter ($\phi$) to provide a weighted interpretation suitable for left-skewed crash data. Higher $\phi$ values reflect the improved capability of the model to represent scenarios with lower risk accurately. Additionally, the index parameter ($\rho$) is essential to capture the long-tail distribution of crash risk values. Values near 2 for $\rho$ indicate the robust performance of the model in these scenarios.

\subsection{The framework of Spatiotemporal Zero-Inflated Tweedie Graph Neural Networks (STZITD-GNNs)} \label{Framework of STGNNs}

In relation to the four parameters $\pi,\mu, \phi, \rho$ of the ZITD probability distributions detailed above, we have designed a probabilistic spatio-temporal graph learning module, named STZITD-GNNs. This module extracts the spatiotemporal correlations $\mathcal{Z}=\mathcal{ST}_{\Theta}(X_{1:t}, Y_{1:t}, A)$ under the assumption of ZITD for each road crash and predicts future results for the subsequent $p$ time windows through $f_{ZITD}(X_{t+1:t+p}|\mathcal{Z})$, as described in Equation \ref{eq:zitd_st}. 
\begin{equation}
    \label{eq:zitd_st}
    \begin{aligned}
        &f_{ZITD}(Y_{t+1:t+p}|\pi_{t+1:t+p},\mu_{t+1:t+p},\phi_{t+1:t+p}, \rho_{t+1:t+p}) \\
        &= f_{ZITD}(Y_{t+1:t+p}|\mathcal{ST}_{\Theta}(X_{1:t},Y_{1:t},A)) = f_{ZITD}(X_{t+1:t+p}|\mathcal{Z}).
    \end{aligned}
\end{equation}

As illustrated in Figure \ref{fig:method}, instead of performing two independent estimations, our methodology starts with the deployment of a GRU acting as a temporal encoder. Subsequently, a GAT is implemented for the purpose of spatial encoding. The combined output from the GRU and GAT is utilised to parameterize the ZITD distribution. The proposed approach is shown in Equation \ref{eq:parameter}. The architecture's encoding component applies a spatio-temporal embedding scheme complemented with an extra sparsity parameter and distribution parameters. The decoding component is responsible for probabilistic estimations of future multi-step crash risk values, which can be construed as the set of parameters of the outcome distribution. In contrast to the principles of variational autoencoders, which restrict latent embedding with a specific Gaussian distribution for the representation of latent variables \citep{kipf2016variational,hamilton2020graph}, we echo the statistical domain by allowing integration of deep learning methodologies into parameter learning.
\begin{equation}
\label{eq:parameter}
    \pi_{t+1:t+p}, \mu_{t+1:t+p}, \phi_{t+1:t+p}, \rho_{t+1:t+p} = \mathcal{ST}_{\Theta}(X_{1:t},Y_{1:t},A).
\end{equation} 

\subsubsection{Parameter Decoders}
The detailed functions of the temporal encoder, GRU and the spatial encoder, GAT are listed in the \ref{app:GRU} and \ref{app:GAT}. In this section, we highlight the parameter decoders in detail. To transmute the acquired spatio-temporal embeddings into ZITD parameter values, we have devised four parameter decoders to operate on $\mathcal{Z}$. Consequently, four parameters, namely $\pi_{t+1:t+p}, \mu_{t+1:t+p}, \phi_{t+1:t+p},$ and $\rho_{t+1:t+p}$ can be computed from $\mathcal{Z}$, as detailed below:
\begin{equation}
\centering
 \begin{aligned}
\pi_{t+1:t+p} &= \sigma(W_{\pi}\cdot \mathcal{Z} + b_{\pi}) \\
\mu_{t+1:t+p} &= \text{ReLU}(W_{\mu}\cdot \mathcal{Z} + b_{\mu}) \\
\phi_{t+1:t+p} &= \text{ReLU}(W_{\phi}\cdot \mathcal{Z} + b_{\phi}) + \epsilon \\
\rho_{t+1:t+p} &= \sigma(W_{\rho}\cdot \mathcal{Z} + b_{\rho}) + 1 + \epsilon
    \label{eq:four parameters at last layer}
 \end{aligned}
\end{equation}
Here, $\pi$ lies within the range of $[0,1]$, $\mu$ falls within the interval $[0, +\infty)$, $\phi$ exists within $(0,+\infty)$, and $\rho$ spans the range of $(1,2)$. The learnable weight matrices are represented as $W_{\pi}, W_{\mu}, W_{\phi}, W_{\rho}, b_{\pi}, b_{\mu}, b_{\phi}, b_{\rho} \in \mathbb{R}^{F' \times p}$. $\lim \epsilon \to 0$ represents the smallest value, and $\text{ReLU}(\cdot)$ signifies the ReLU activation function.

\subsection{Learning Framework of STZITD-GNNs}
\label{3.4}
\subsubsection{ZITD Loss Function}
As previously mentioned, our encoding mechanism employs a spatio-temporal embedding architecture complemented by a sparsity parameter. The decoding component involves a probabilistic estimation of future road traffic crashes. To accurately predict these crashes, the overall learning objective can be seen as maximising the log-likelihood function: $\max \log f_{ZITD}(y_k|\pi,\mu, \phi, \rho)$. We use the negative likelihood as our loss function to better fit the distribution into the data. Here, $y$ denotes the ground-truth values corresponding to one of the predicted crashes with the parameters $\pi,\mu, \phi, \rho$. The log-likelihood of ZITD is made up of the parts $y=0$ and $y>0$:

For $y>0$, $NLL_{y>0}=-\log f_{ZITD}(y>0|\pi,\mu, \phi, \rho)$:
\begin{equation}
\begin{aligned}
    \log f_{ZITD}(y>0|\pi,\mu, \phi, \rho) &=
    \log(1-\pi) + \log f_{\text{TD}}(y>0|\mu, \phi, \rho)   \\
    &= \log(1-\pi) + \frac{1}{\phi} \bigl (y \frac{\mu ^{1-\rho}}{1-\rho} - \frac{\mu ^{2-\rho}}{2-\rho}  \bigl) + \log a(y>0, \phi, \rho) \\
    & = \log(1-\pi) + \frac{1}{\phi} \bigl (y \frac{\mu ^{1-\rho}}{1-\rho} - \frac{\mu ^{2-\rho}}{2-\rho}  \bigl) - \log y + \log \sum_{j=1}^{\infty} 
        \frac{y^{-j \alpha}(\rho-1)^{\alpha j} }{\phi^{j(1-\alpha)} (2-\rho)^j j! \Gamma(-j\alpha)} \\  
    & \geq \log(1-\pi) + \frac{1}{\phi} \bigl (y \frac{\mu ^{1-\rho}}{1-\rho} - \frac{\mu ^{2-\rho}}{2-\rho}  \bigl) - \log (j_{max} \sqrt{-\alpha} y ) + j_{max}(\alpha - 1). 
\end{aligned}
\end{equation}
In this equation, $j_{max}=\frac{y^{2-\rho}}{(2-\rho)\phi}$ and $\alpha = \frac{2-\rho}{1-\rho}<0$. We optimise the lower bound; hence, $NLL_{y>0}$ can be optimised in this manner.
    
For $y=0$, $NLL_{y=0}=-\log f_{ZITD}(y=0|\pi,\mu, \phi, \rho)$:
\begin{equation}
\begin{aligned}
    \log f_{ZITD}(y=0|\pi,\mu, \phi, \rho) &=
    \log(\pi) + \log(1-\pi)  f_{\text{TD}}(y=0|\mu, \phi, \rho) \\
    &= \log(\pi) + \log(1-\pi) +  \frac{1}{\phi} \bigl ( - \frac{\mu ^{2-\rho}}{2-\rho}  \bigl)
\end{aligned}
\end{equation}
\noindent Here, $\pi,\mu, \phi, \rho$ are chosen and calculated in accordance with the index of $y=0$ or $y>0$. The ultimate negative log-likelihood loss function is given by:
\begin{equation}
    NLL_{STZITD} = NLL_{y=0} + NLL_{y>0} + \eta{\Theta^2}.
    \label{eq:nll_all}
\end{equation}
Model parameter $\Theta$ can be optimized by minimizing the negative log-likelihood loss: $\hat{\Theta} = \text{argmin} NLL_{STZITD}$. $\eta$ stands for the weight parameter for $L_2$ Normalisation.

It is worth noting that our model can be extended to other distributions by modifying the probability layer. The main reason is that we learn the parameters that determine the distributions. For example, if we choose the Gaussian distribution, we can parameterize the probability layer using the spatio-temporal embedding of the mean and variance, thereby quantifying the data uncertainty that follows the Gaussian distribution. Using the flexibility of the probability layer, we design a range of benchmark models for comparison to the STZITD-GNNs. Our code is available on Github \footnote{\url{https://github.com/STTDAnonymous/STTD}}.

\subsubsection{Time analysis \& Training algorithms} 
We summarise the STZITD-GNN training procedure in Algorithm \ref{alg:training}. Based on that, the time complexity can be analyzed as follows (for the sake of simplicity, we denote the hidden union of embedding $F'\approx F$ and $F$ as $F$):
\begin{itemize}[leftmargin=*]
    \item For temporal encoders, the time complexity can be formulated as $O\bigl(t N(dF+F^2+F) \bigl)$. 
    \item For spatial encoders, we assume that the number of road edges is $|E|$. Hence, when the graph attention encoder is implemented, the complexity of computing attention becomes $O(M|E|F)$. Furthermore, when performing $W_a$ multiplication, the cost is $O(M N F^2)$. This leads to a total time complexity of $O(M|E|F+M N F^2)$.
    \item When generating the four parameters, the four one-layer MLPs require a time complexity of $O\bigl(F^2 p\bigl)$.
\end{itemize}

\begin{algorithm}[H]
    \caption{The training of STZITD-GNNs.}
    \small
    \label{alg:training}
    \renewcommand{\algorithmicensure}{\textbf{Output:}}
    \begin{algorithmic}[1]
    \Require Spatiotemporal road feature $X \in \mathbb{R}^{N \times T \times d}$, road network graph $G=(V, E, A)$, roads traffic crash $Y \in \mathbb{R}^{N \times T}$, model hyperparameters $M, F', F$, and total time slot $T$, time window $t$, $p$. 
    \State Calculate road traffic crash $Y$ via Eq.~\eqref{eq: y_it}.   
    \State Initialise parameters $\Theta$ via Xavier Initializer.
    \While{STZITD-GNNs is not converged}     \Comment{Train}
        \State Calculate temporal embedding $\mathcal{Z}_T$ via Eq.~\eqref{eq:GRU}  by road feature $X_{1:t}$;   \Comment{Temporal Encode}
        \State Calculate spatiotemporal embedding $\mathcal{Z}$ via Eq.~\eqref{eq:multi-head last layer} by graph $G$ and $\mathcal{Z}_T$;   \Comment{Spatial Encode}
    \State Obtain four parameters $\pi_{t+1:t+p}, \mu_{t+1:t+p}, \phi_{t+1:t+p}, \rho_{t+1:t+p}$ via output layer refer to Eq.~\eqref{eq:four parameters at last layer};
    \State Calculate distribution $f_{ZITD}(y_{t+1:t+p}|\pi_{t+1:t+p}, \mu_{t+1:t+p}, \phi_{t+1:t+p}, \rho_{t+1:t+p})$ from four parameters via Eq.~\eqref{eq:zitd_den};
    \State Minimizing negative log-likelihood loss ${NLL}_{STZITD}$ against 
 $Y_{t+1:t+p}$ via Eq.~\eqref{eq:nll_all} using Adam Optimizer;
    \EndWhile
    \State End optimizing model parameters $\Theta$; 
    \end{algorithmic}
\end{algorithm}

\section{Experiments}
\label{sec:experiements}
In this section, we describe the detailed elements of the STZITD-GNNs model by analysing its performance in crash prediction in three boroughs of London, UK, namely Lambeth, Tower Hamlets and Westminster, each characterised by distinct sociodemographic traits, road networks, and traffic crash profiles. The crash data is accessed from the UK STATS19 dataset, maintained by the Department for Transport. It is made up of three interlinked tables that provide information about crashes, their severity, and related vehicle data \footnote{https://www.gov.uk/guidance/road-crash-and-safety-statistics-guidance}. We have configured the model to forecast 14 steps ahead, achieving stable temporal horizon results. Our evaluation focusses on two key aspects: first, demonstrating the model’s ability to effectively quantify the uncertainty of road-level traffic crash data, and second, showing its precision in predicting. We emphasise that our model not only forecasts multi-step crash risk in advance but also does so with a higher degree of reliability and precision at the finer granularity compared to several relevant baseline models.

Section~\ref{sec:data description} introduces the datasets used in this study and justifies the choice of data-driven strategies. This is followed by an outline of the experimental setup in Section~\ref{experiment setup}. Subsequently, we describe the evaluation metrics in Section~\ref{sec:evaluation metrics} and the baseline models used for comparison in Section~\ref{sec: baseline models}. The subsequent analysis includes a comparison of the performance of our model against other methods in Section~\ref{sec: performance comparison}. It concludes with a sensitivity analysis in Section~\ref{sensitive analysis}

\subsection{Traffic Crash Data Description}
\label{sec:data description}


\begin{figure}[!htbp]
\vspace{-0.5mm}
    \begin{minipage}{0.51\columnwidth}
	\centering
        \includegraphics[width=0.69\columnwidth]{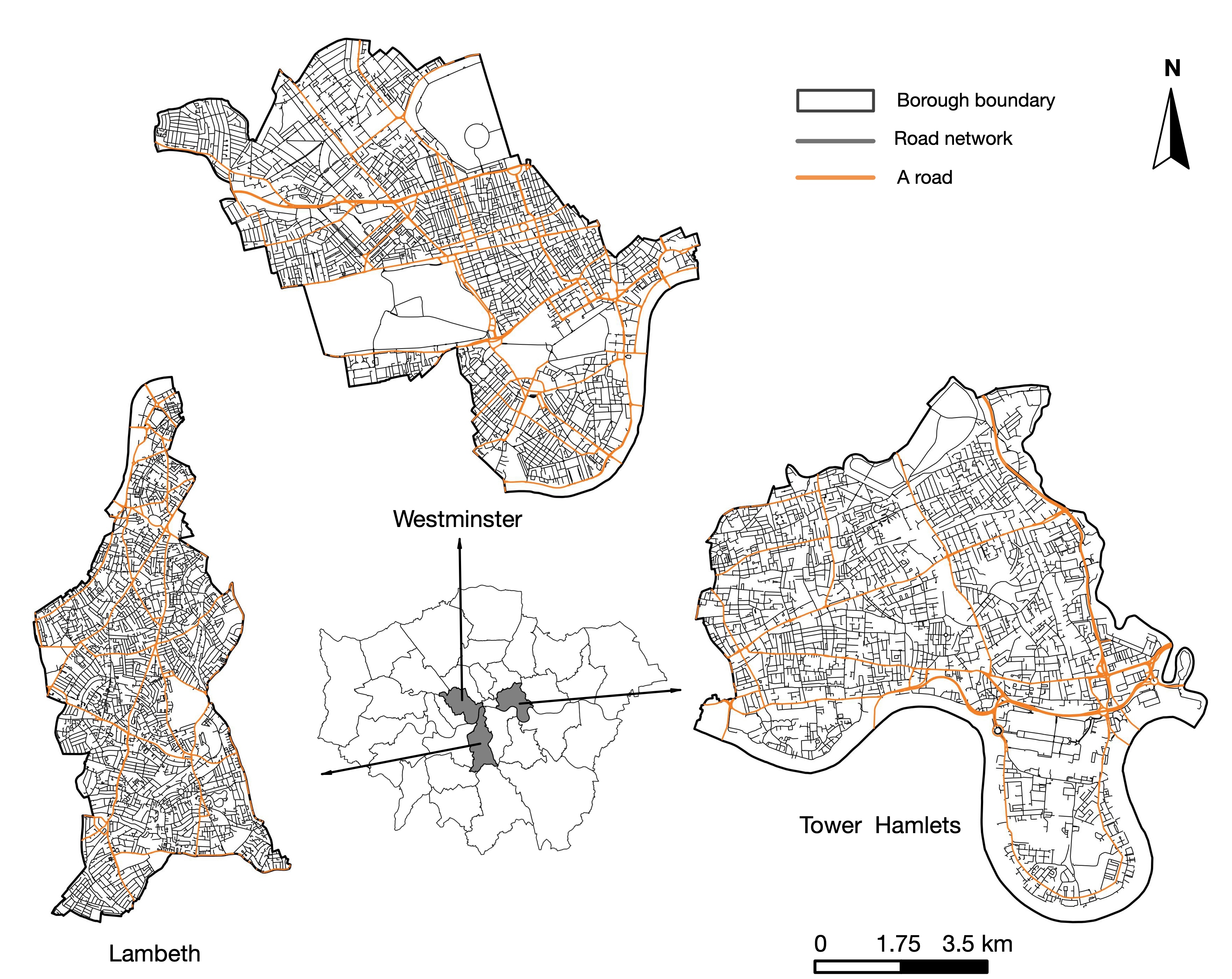}
	\subcaption{The map of three boroughs, London}
        \label{fig:map_three}
	\end{minipage}
    \begin{minipage}{0.5\columnwidth}
	\centering
        \includegraphics[width=0.99\columnwidth]{ 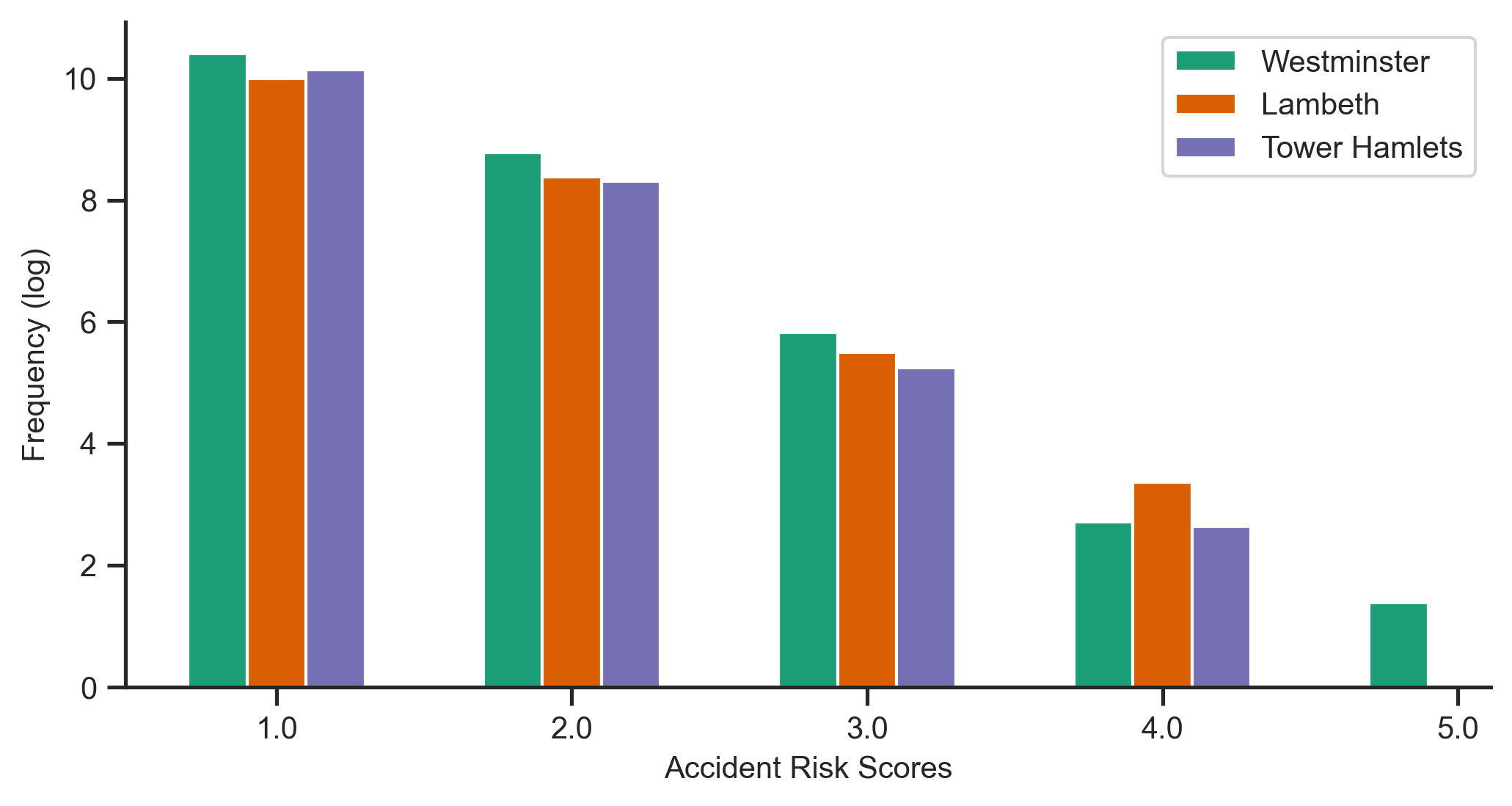}
	\subcaption{The distribution of 2019-crash risk scores frequency in the three boroughs}
        \label{fig:all_risk}
   \end{minipage}
 
	  \caption{Traffic crashes are depicted for the case studies.}
\label{fig:mds_vis_risk.png}
\end{figure}

The three high-crash frequency boroughs of Greater London, UK are illustrated in Figure \ref{fig:map_three}. These boroughs each averaged 3 to 5 daily crashes throughout 2019, as detailed in Table \ref{tb:data}. Despite these figures, a significant number of roads in these boroughs report no crashes, leading to a pronounced zero-inflation problem. This situation is compounded by the distribution of crash risk values (Figure \ref{fig:all_risk}), which shows a zero-inflated problem along with a pronounced long-tailed effect -where a few high-risk crash occurrences contrast with many low-risk ones. Such data characteristics underscore the suitability of our proposed model for this dataset, effectively capturing both the prevalent zero rates and the diverse spread of crash risk levels.

\begin{table*}[htb]
  \centering
  \setlength{\abovecaptionskip}{0.1cm}
  \caption{The statistics of Westminster, Lambeth, and Tower Hamlets datasets in 2019.}
    \begin{threeparttable}
\begin{tabular}{c|cccccc}
    \hline 
     \mbox{}\quad\textbf{Dataset}\quad\mbox{} & \mbox{}~\textbf{Roads/Nodes}~\mbox{} &  \mbox{}~\textbf{Edges}~\mbox{} & \mbox{}~\textbf{Zero-Inflated}~\mbox{}\\ \hline
     Westminster${}^\text{1}$ & 4,822 & 20,128 & 95.72\% \\
     Lambeth${}^\text{1}$ & 5,659 & 21,574 & 96.71\% \\
     Tower Hamlets${}^\text{1}$  & 4,688 & 17,928 & 96.28\% \\ \hline
\end{tabular}
    \begin{tablenotes}
      \footnotesize
      \item ${}^\text{1}$ {https://roadtraffic.dft.gov.uk/downloads}
     \end{tablenotes}
    \end{threeparttable}
  \label{tb:data}%
  \vspace{-0.15cm}
\end{table*}

\subsection{Experiment Setup}
\label{experiment setup}
The train, validation and test data are all from 2019, as the ratio is 8:2:2. The parameters of STZITD-GNN are optimised with the Adam optimiser \citep{kingma2014method}, where the learning rate is set to 0.01, and weight decay is 0.01. In STZITD-GNN, $N_{epoch}=20$, the hidden unit is set to 42, GAT head is set to 3. The baseline parameters are optimised using Adam \citet{kingma2014method} with regularisation of $L_2$ and a dropout rate of 0.2. The GNNs in STZITD-GNNs and baselines are all two-layered. We also set the hidden units at 42. Similarly to previous work, we also employ the early stopping strategy with patience equal to 10 to avoid overfitting \citep{zhou2020accident}. STZITD-GNNs is implemented in Pytorch 1.9.0 with Python 3.8. All experiments were carried out on 1 NVIDIA GeForce RTX 3090, 24 GB. 

\subsection{Evaluation Metrics}
\label{sec:evaluation metrics}
We use eight metrics to evaluate the performance of our STZITD-GNNs from four perspectives, including the predictive regression effect of road crash risk scores, the predictive effect of uncertainty-aware distributional characteristics, the accuracy of zero crash road identification, and the accurate identification of crashes occurring on predicted high-risk roads.

\subsubsection{Crash Accuracy Metrics}
For the aspect of the point estimate, we compare the prediction of the numerical crash risk scores. For probabilistic distribution outputs like our model, we use the mean value of the distribution as the numerical output. For specific metrics, we use Mean Absolute Error (MAE), Mean Absolute Percentage error (MAPE), and Root Mean Squared Error (RMSE) to evaluate the predicted crash risk scores of each road. Specific definitions are summarized:
\begin{equation}
\centering
 \begin{aligned}
&\text{MAE}=\frac{1}{pN} \sum_{j=1}^{p} \sum_{i=1}^{N}  \left|y_{ij} - \hat{y}_{ij}\right|, 
\\
& \text{MAPE} = \frac{1}{pN} \sum_{j=1}^{p} \sum_{i=1}^{N}  \left|
\frac{y_{ij} - \hat{y}_{ij}}{y_{ij}} \right |,
\\ 
& \text{RMSE} = \sqrt{\frac{1}{pN}\sum_{j=1}^{p} \sum_{i=1}^{N} (y_{ij} - \hat{y}_{ij})^2},
\label{eq:metric 1}
\end{aligned}
\end{equation}
\noindent where $\hat{y}_{ij}$ and $y_{ij}$ are the predicted (i.e. distribution mean) and ground-truth values of $i^{th}$ road at time slot $j^{th}$ respectively. $p= 14$ days, and the time interval $j^{th}$ is the predicted window index. Lower values of these metrics mean better performance with fewer errors.

\subsubsection{Uncertainty Quantification Metric}\label{sec:metric}
To quantify the uncertainty of the results, we use two typical metrics: the mean prediction interval width (MPIW) and the prediction interval coverage probability (PICP) are based on a confidence interval 5\%-95\% with a confidence level set at 5\% ($\alpha=5\%$).
MPIW gauges the average width of the prediction interval, reflecting the distribution's specific characteristics. Ideally, a smaller MPIW indicates reduced variance and increased certainty, highlighting a more precise prediction. The PICP, on the other hand, is implemented to evaluate whether the ground truth value, denoted $y_{ij}$, lies within the predicted interval. A higher PICP indicates better performance. The goal is to achieve a prediction interval that is narrow and encompasses a significant number of data points. 

\begin{equation}
\centering
 \begin{aligned}
& \text{MPIW}=\frac{1}{pN}\sum_{j=1}^{p} \sum_{i=1}^{N}  (U_{ij} - L_{ij}), \\
& \text{PICP}=\frac{1}{pN}\sum_{j=1}^{p} \sum_{i=1}^{N}  \mathcal{I}(U_{ij}< \hat{y}_{ij} <L_{ij}),
\label{eq:metric 2}
\end{aligned}
\end{equation}
$L_{ij}$ and $U_{ij}$ correspond to the lower and upper bound of the confidence interval for observation road $i$ at time slot $j^{th}$. The $\mathcal{I}(U_{ij}< \hat{y}_{ij} <L_{ij})$ equals 1 if the condition is true, else 0. 

\subsubsection{Zero crash Identification Metrics}
Given the problem of zero-value data sparsity, highlighted in Figure \ref{fig:all_risk}, it is essential to assess the ability of a model to accurately represent roads without crashes. We used the true zero rate (ZR) as a metric to quantify whether a model reflects the zero sparsity seen in the actual data. The formula is as follows: 
\begin{equation}
\centering
 \begin{aligned}
& \text{ZR} = {\frac{1}{pN}\sum_{j=1}^{p} \sum_{i=1}^{N} \mathcal{I}(y_{ij}=0 \cap \hat{y}_{ij}=0)},
\label{eq:metric 3}
\end{aligned}
\end{equation}
where the indicator function, $\mathcal{I}(y{ij}=0 \cap \hat{y}{ij}=0)$, equals 1 if the condition holds true, and 0 otherwise. 

\subsubsection{Accuracy Hit Rate Metric on Crash Occurence} 

To evaluate the effectiveness of our model in identifying roads with a higher likelihood of crash occurrence, we use the accuracy hit rate in $a$ ($AccHR@20$) metric. Specifically, we focus on the top 20\% ($a=20\%$) of roads considered high-risk to verify if these areas actually correspond to actual crash occurrence. This alignment of predicted high-risk areas with real crash sites, as reported in \citet{zhang2020graph,zhang2019graph}, allows Acc@$a$ to effectively measure the precision of our model in forecasting roads at increased risk of crashes. The formula for calculating Acc@$a$ is as follows:

\begin{equation}
\centering
 \begin{aligned}
AccHR@a  = \frac{1}{p}  \sum_{j=1}^{p} \frac{\sum_{i=1}^{N} \mathcal{I}(y_{ij}>0 \cap \hat{y}_{ij} \geq \text{Top}(\hat{y}_{:j}, a) )}{\sum_{i=1}^{N}\mathcal{I}(y_{ij}>0)},
\label{eq:metric}
\end{aligned}
\end{equation}

The indicator function, $\mathcal{I}(y_{ij}>0 \cap \hat{y}{ij} \geq \text{Top}(\hat{y}{:j}, a))$, is rated as 1 when a crash (denoted as $y_{ij}>0$ for a crash occurrence on the $i^{th}$ road in the $j^{th}$ time interval) aligns with the top $a$ percent of predicted risks (where $\hat{y}{ij}$ meets or exceeds the threshold $\text{Top}(\hat{y}{:j}, a)$), and 0 otherwise.

\subsection{Baseline Models}
\label{sec: baseline models}
In evaluating our proposed model, we carefully select baseline models that allow evaluation of the features of the proposed model: uncertainty quantification in multistep road-level crash risk prediction. Our selection criteria favour traditional point estimation methods and the most recent interval prediction models. This approach is in line with recent studies \citet{zhuang2022uncertainty}. Specifically, our baselines include the mean-predict method (HA) and advanced spatial-temporal graph deep learning techniques such as STGCN and STGAT. Additionally, we consider parameterisation approaches such as STG-GNN, STNB-GNN, STTD-GNN, and STZINB-GNN. These probabilistic models have two or three parameter distributions, simpler than the four-parameter models we propose. The comparison models are:
\begin{itemize}
\item \textbf{HA}(\cite{liu2004summary}): The historical average serves as the minimum baseline. It is calculated by averaging road traffic crashes in the same time intervals from historical data to predict the future road traffic crash. 
\item \textbf{STGCN}(\cite{STGCN}): Spatio-temporal graph convolution network, which utilises graph convolution and 1D convolution to capture spatial dependencies and temporal correlations, respectively. 
\item \textbf{STGAT}(\cite{GAT}): The Graph Attention Network (GAT) can focus on the features of neighbouring nodes by assigning different attention weights to each node within a neighbourhood. We replaced the GCN with GAT in the STGCN and named the modified model STGAT.

\item \textbf{STG-GNN}(\cite{wang2023uncertainty}): Same as STZITD-GNN, we replace the statistical assumption of road traffic crashes with Gaussian distribution.
\item \textbf{STNB-GNN}(\cite{zhuang2022uncertainty}): Same as STZITD-GNN, we replace the statistical assumption of road traffic crashes with Negative Binomial Distribution.
\item \textbf{STTD-GNN}: Same as STZITD-GNN, we replace the statistical assumption of road traffic risks with TD Distribution.
\item \textbf{STZINB-GNN}(\cite{zhuang2022uncertainty}): Unlike STZITD-GNN, we replace the statistical assumption of road traffic crashes with a zero-inflated negative binomial distribution.
\end{itemize}

\subsection{Performance Comparison}
\label{sec: performance comparison}

We evaluated the performance of our proposed model, STZITD-GNN, against several baseline models from four different aspects: accuracy of point estimation, uncertainty quantification, non-risk road identification and accurate prediction of crash occurrence, as shown in Tables \ref{tab:long}. The results are indicated by upward arrows, representing improvements over the second-best models. These enhancements confirm the significant impact of probabilistic assumptions on the accuracy and stability of our results, consistent with recent findings by \citet{wang2023uncertainty}. The consistent performance of the STZITD-GNN model underscores the critical importance of selecting appropriate distributional assumptions that align with the characteristics of traffic crash data. 

For point estimation accuracy, the superior performance of the STZITD-GNN model across all regression error metrics highlights the effectiveness of the ZITD distribution. This distribution captures the long-tail dynamics of road traffic crash risk data with greater accuracy, making the mean value derived from the four parameters of the distribution decoder more representative of the actual risk scores. STZITD-GNN significantly improves the accuracy of the estimation of the crash risk score, demonstrating a 49.06\% increase in MAPE compared to the second-best model in the Lambeth case study, which has the most significant zero-inflation. Traditional deterministic models like HA, STGCN, and STGAT show varied performance, with HA consistently underperforming. GAT-based models, utilising predefined adjacency matrices that embed domain spatial knowledge, generally outperform GCN models by incorporating spatial proximity and contextual similarity, thereby better representing spatial dependencies. Among probabilistic models, those that assume Gaussian distributions perform poorly. They are significantly outperformed by deterministic models and show far worse performance 50\% compared to the STZITD-GNN model, which uses skewed distribution assumptions. This underscores the importance of selecting appropriate distributional assumptions for probabilistic models. 

Second, the variation in uncertainty quantification across different models highlights the significant impact of distributional assumptions.
The four-parameter distribution model (STZITD-GNN) shows substantial improvements, most notably in effectively capturing a narrow distribution while encompassing a wider range of actual crash risk values. In contrast, the worst-performing model, STG-GNNs, illustrates how improper distributional assumptions can degrade the quality of the estimation, introducing greater uncertainty into the results. The MPIW of this model is approximately 1.5 times that of the least uncertain three-parameter model, due to its constraints that enforce equal mean and variance in the distribution. Although ideally, outcome distributions would have a narrow interval, traditionally a wider MPIW suggests a higher PICP, indicating a trade-off in measuring uncertainty \citep{wang2023uncertainty,zhu2022cross,zhuang2022uncertainty}. The STZINB-GNN and STTD-GNN models show that in scenarios characterised by extreme zero-inflation and heavier tails, a higher MPIW is associated with better PICP performance. This correlation suggests that broader prediction intervals, which encompass more observations, are beneficial in more extreme cases as they cover a wider range of observed values, despite reducing the ability to achieve narrower intervals. However, the STZITD-GNN model shows a promising improvement in MPIW, ranging from 47.30\% to 55.07\% in all case studies, highlighting its efficacy. In terms of PICP, it also shows a marginal increase of approximately 0.2\% over the next best model, STTD-GNNs. This modest improvement still validates the impact of leveraging long-tailed and zero-inflated TD distributions, which contribute to a larger PICP and reduced variance in the predicted outcomes.

In scenarios characterised by a high prevalence of non-crash roads, the STZITD-GNN model not only aims to precisely quantify crash risk values but also effectively differentiates non-crash roads. The integration of the sparsity parameter ($\pi$) significantly enhances the ability of the traditional TD assumption to address the prevalent zeros in the dataset, thereby reducing uncertainties associated with zero outcomes. The TD and ZINB distributions provide distinct parametric approaches to tackle the issue of zero-inflation. The TD distribution naturally integrates zero-inflation through its end-to-end structure, while the ZI parameter in the ZINB model requires an additional interpretative step like ZITD model does. For example, in the Tower Hamlets and Westminster cases, the ZI parameter significantly increases performance in zero-rate predictions compared to the NB distribution, showing improvements of 16.54\% and 17.94\%, respectively. In extreme zero-inflation scenarios, such as in Lambeth, both the STTD and STZINB models perform poorly, with a 15.09\% discrepancy in ZR results compared to the STZITD-GNN model. Although the ZI paradigm is integrated into both the TD and the NB models, the ZITD model has a continuous parameter space, which can be advantageous for fitting data with a mix of zero inflation and continuous positive values. In contrast, traditional ZINB models operate under a discrete probabilistic assumption and struggle with complex data structures, particularly when the positive values are continuous rather than merely counts.

In analyses that correlate high-risk roads with actual crashes, the STZITD-GNN model consistently demonstrates robust performance. In particular, it achieved a significant 76.59\% precision in identifying actual crashes in the Lambeth case. Across all three distinct cases, the model maintains a robust and accurate identification rate, with a variance of only 7.61\%. In contrast, models employing unsuitable distribution assumptions, such as STG-GNN, tend to significantly underpredict crash occurrences, often yielding the poorest results. This discrepancy underscores the importance of selecting the appropriate distribution model again.

\begin{table*}[!ht]
\caption{Comparison of multi-step prediction performance (14 days) in three boroughs. The bold font indicates the best performing model, while underline indicates the second best. To maintain clarity, detailed names of all GNNs in the final five statistically integrated deep learning models are omitted. Instead, abbreviations such as STG, STNB, STTD, STZINB, and STZITD are used to emphasise the most significant modifications within the GNN models.}
\small
\resizebox{\linewidth}{!}{
\begin{tabular}{m{1.8cm}<{\centering}m{2cm}<{\centering}|m{1cm}<{\centering}m{1cm}<{\centering}m{1cm}<{\centering}m{1cm}<{\centering}m{1cm}<{\centering}m{1cm}<{\centering}m{1cm}<{\centering}m{1.2cm}<{\centering}}
\toprule
\multicolumn{1}{c|}{Metric} & Model & HA & STGCN & STGAT & STG & STNB & STTD & STZINB & STZITD \\
\midrule
\multicolumn{2}{c|}{Dataset} & \multicolumn{8}{c}{Lambeth} \\
\midrule
\multicolumn{1}{c|}{\multirow{3}{*}{\makecell[c]{Point\\ Estimation}}} & MAE & 0.1348 & 0.0613 & \underline{0.0302} & 0.1182 & 0.0802 & 0.0315 & 0.0544 &  $\mathbf{0.0238}_{\uparrow21.19\%}$ \\
\multicolumn{1}{c|}{} & MAPE & 0.4142 & 0.4023 & 0.3203 & 0.4046 & 0.0363 & 0.0426 & \underline{0.0265} &  $\mathbf{0.0135}_{\uparrow49.06\%}$ \\
\multicolumn{1}{c|}{} & RMSE & 0.2107 & 0.1436 & 0.1122 & 0.1827 & 0.1390  & \underline{0.1098} & 0.1191 &  $\mathbf{0.0947}_{\uparrow13.75\%}$ \\ \cline{1-2} 
\multicolumn{1}{c|}{\multirow{2}{*}{\makecell[c]{Uncertainty\\ Quantification}}} & MPIW & / & / & / & 0.4871 & 0.1946 & 0.1349 & \underline{0.0454} & $\mathbf{0.0204}_{\uparrow55.07\%}$ \\
\multicolumn{1}{c|}{} & PICP & / & / & / & 0.7328 & 0.8903 & \underline{0.9873} & 0.9796 & $\mathbf{0.9899}_{\uparrow00.26\%}$ \\ \cline{1-2}
\multicolumn{1}{c|}{\makecell[c]{Non crash \\ Quantification}} & ZR & 0.5203 & 0.5305 & 0.5986 & 0.1992 & 0.5621 &\underline{0.6838} & 0.6408 &  $\mathbf{0.7870}_{\uparrow15.09\%}$ \\ \cline{1-2}
\multicolumn{1}{c|}{Hit Rate} & $AccHR@20$ & 0.4520 & 0.6113 & 0.6422 & 0.2666 & 0.4471 & \underline{0.7123} & 0.6184 &  $\mathbf{0.7659}_{\uparrow07.52\%}$ \\
\midrule
\multicolumn{2}{c|}{Dataset} & \multicolumn{8}{c}{Tower Hamlets} \\
\midrule
\multicolumn{1}{c|}{\multirow{3}{*}{\makecell[c]{Point\\ Estimation}}} & MAE & 0.0939 & 0.2385 & \underline{0.0305} & 0.1741 & 0.2475 & 0.0433 & 0.2135 &  $\mathbf{0.0271}_{\uparrow11.15\%}$ \\
\multicolumn{1}{c|}{} & MAPE & 0.2896 & 0.2017 & 0.1205 & 0.2662 & 0.0399 & \underline{0.0246} & 0.0594 &  $\mathbf{0.0221}_{\uparrow10.16\%}$ \\
\multicolumn{1}{c|}{} & RMSE & 0.1987 & 0.1325 & 0.1355 & 0.3896 & 0.1548 & \underline{0.1186} & 0.1368 &  $\mathbf{0.0918}_{\uparrow22.60\%}$ \\ \cline{1-2} 
\multicolumn{1}{c|}{\multirow{2}{*}{\makecell[c]{Uncertainty\\ Quantification}}} & MPIW & / & / & / & 0.5321 & 0.2263 & 0.0810 & \underline{0.0296} &  $\mathbf{0.0156}_{\uparrow47.30\%}$ \\
\multicolumn{1}{c|}{} & PICP & / & / & / & 0.6439 & 0.9442 & \underline{0.9845} & 0.9845 &  $\mathbf{0.9871}_{\uparrow00.26\%}$ \\ \cline{1-2} 
\multicolumn{1}{c|}{\makecell[c]{Non crash \\ Quantification}} & ZR & 0.4998 & 0.4896 & 0.5234 & 0.1351 & 0.5158 & 0.6420  & \underline{0.7526} &  $\mathbf{0.8876}_{\uparrow17.94\%}$ \\ \cline{1-2} 
\multicolumn{1}{c|}{Hit Rate} & $AccHR@20$ & $0.4752$ & 0.5869 & \underline{0.6950}  & 0.2647 & 0.5022 & 0.6368 & 0.5827 &  $\mathbf{0.7224}_{\uparrow03.94\%}$ \\
\midrule
\multicolumn{2}{c|}{Dataset} & \multicolumn{8}{c}{Westminster} \\
\midrule
\multicolumn{1}{c|}{\multirow{3}{*}{\makecell[c]{Point\\ Estimation}}} & MAE & 0.1467 & 0.0784 & 0.0847 & 0.1055 & 0.3266 & \underline{0.0579} & 0.3240  &  $\mathbf{0.0357}_{\uparrow38.34\%}$ \\
\multicolumn{1}{c|}{} & MAPE & 0.1884 & 0.8976 & 0.6012 & 2.9295 & \underline{0.0330}  & 0.0973 & 0.0415 &  $\mathbf{0.0259}_{\uparrow21.52\%}$ \\
\multicolumn{1}{c|}{} & RMSE & 0.5999 & 0.2384 & 0.2433 & 0.6681 & 0.2909 & \underline{0.1552} & 0.2751 &  $\mathbf{0.1015}_{\uparrow34.60\%}$ \\ \cline{1-2} 
\multicolumn{1}{c|}{\multirow{2}{*}{\makecell[c]{Uncertainty\\ Quantification}}} & MPIW & / & / & / & 0.8542 & 0.6272 & 0.2021 & \underline{0.0493} &  $\mathbf{0.0259}_{\uparrow47.46\%}$ \\
\multicolumn{1}{c|}{} & PICP& / & / & / & 0.6036 & 0.9363 & \underline{0.9801} & 0.9796 &  $\mathbf{0.9893}_{\uparrow00.94\%}$ \\ \cline{1-2} 
\multicolumn{1}{c|}{\makecell[c]{Non crash \\ Quantification}} & ZR & 0.5981 & 0.1431 & 0.2922 & 0.2071 & 0.5393 & 0.5079 & \underline{0.6288} &  $\mathbf{0.7328}_{\uparrow16.54\%}$ \\ \cline{1-2} 
\multicolumn{1}{c|}{Hit Rate} & $AccHR@20$ & 0.4217 & 0.5020  & 0.4808 & 0.2933 & 0.4503 & \underline{0.6075} & 0.5139 &  $\mathbf{0.6898}_{\uparrow13.55\%}$ \\
\bottomrule
\end{tabular}
}
\label{tab:long}
\end{table*}

\subsubsection{Visualization of Predicted Road-Level crash}

\begin{figure*}[!htp]
    \centering
    \includegraphics[width=1.05\linewidth]{ 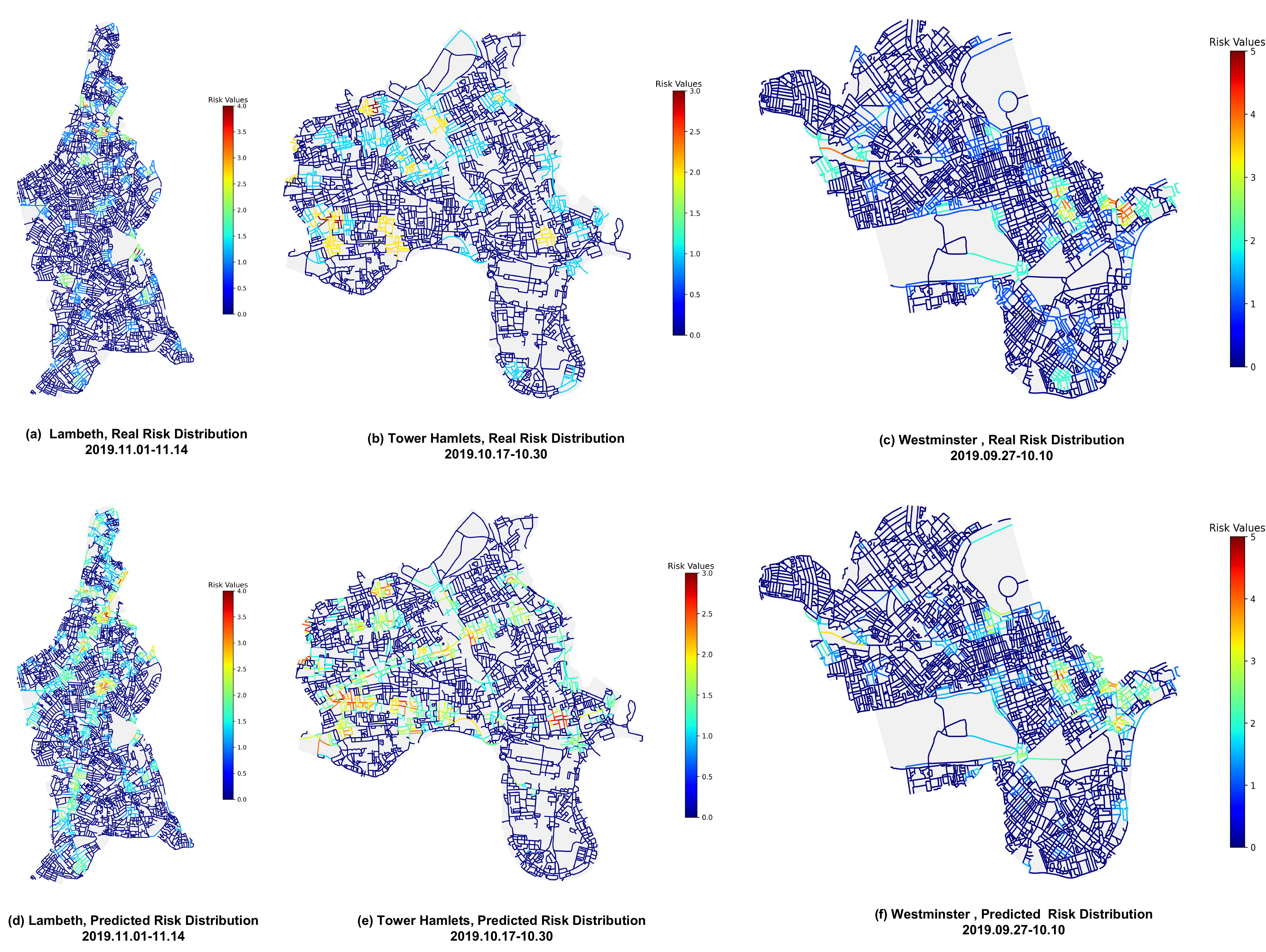}
\caption{Real v.s. Predicted Crash Risks on Lambeth, Tower Hamlets, and Westminster.}
\label{fig:map_long}
\end{figure*}

To illustrate the results, we have generated Choropleth maps of road-level crashes across three distinct boroughs over a 14-day period. Considering the sparsity of crash data, which complicates the visualisation, we selected the three most risk-intensive periods of the data for display, as shown in Figure \ref{fig:map_long}. Each borough reveals distinct crash patterns that correlate closely with the model's predictions, offering a granular view of its predictive accuracy.

These visualizations highlight the general sparsity of traffic crash data, with most roads being non-risk (blue). Only a few roads exhibit higher crash risk levels (red), while others are associated with low-risk crashes. The STZITD-GNN model excels in identifying a range of risk levels, from zero-risk to high-risk. It performs particularly well in Westminster, accurately identifying a broader range of risk-prone roads, aided by the area's lower zero inflation rate and denser graph structure. Furthermore, the model consistently and accurately maps the spatial connectivity of high-risk crash roads across all three boroughs, with risk areas typically clustering along major thoroughfares such as A11 Whitechapel Road and A13 in Tower Hamlets and A3 in Lambeth.

In the three distinct boroughs, the STZITD-GNN model’s ability to handle diverse spatial patterns and variations in daily crashes demonstrates its adaptability and consistent performance. However, the model occasionally misclassifies zero-risk roads as low-risk, particularly in Lambeth and Tower Hamlets. Although these errors generally fall within lower-risk categories and do not significantly impact the identification of higher-risk roads with actual crash occurrences, refining this aspect of the model could further improve its overall accuracy.

\subsection{Sensitivity Analysis}\label{sensitive analysis}
\subsubsection{Temporal Stability of Metrics}

\begin{figure*}[!h]
\centering \includegraphics[width=0.90\linewidth]{ 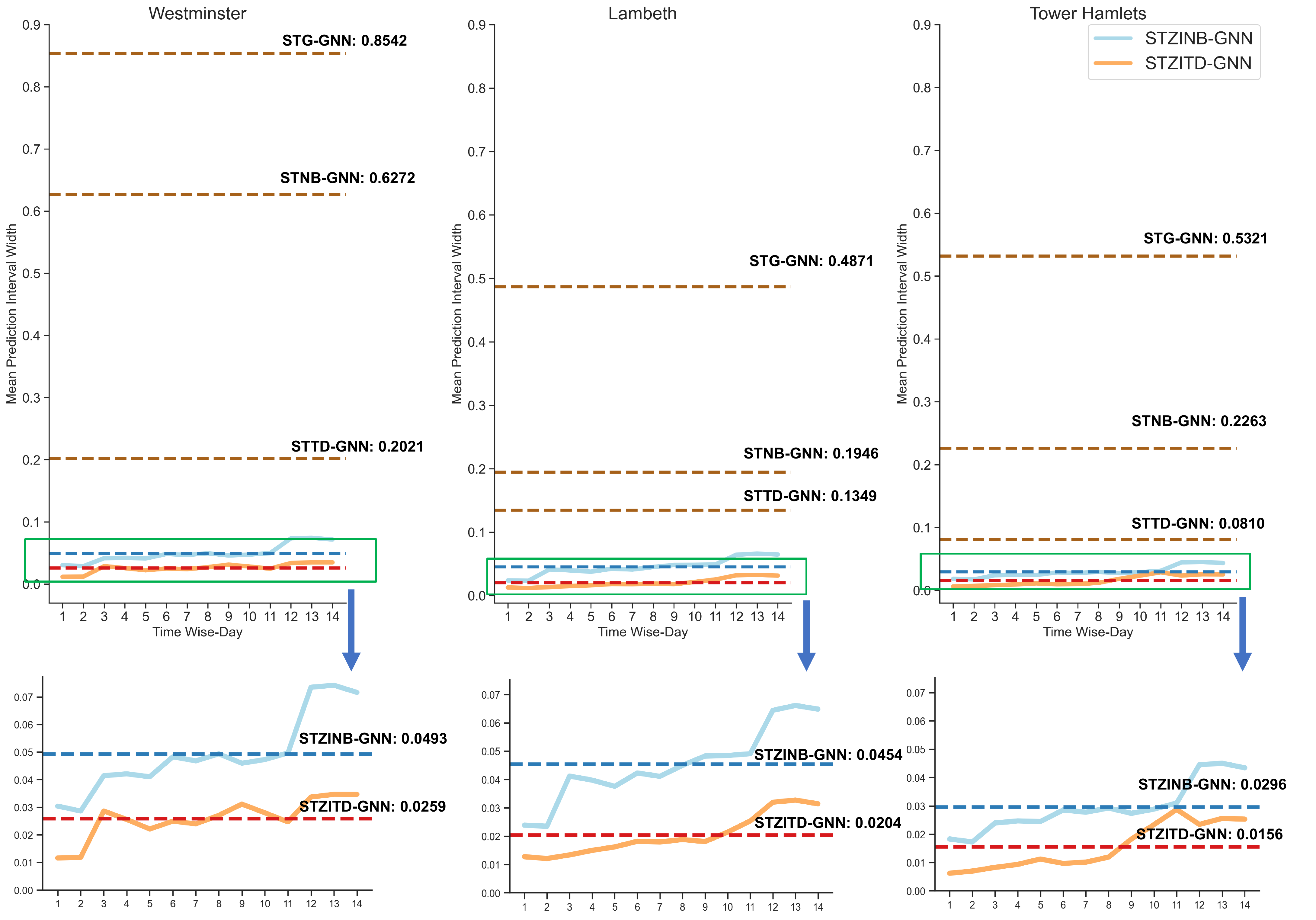}
\caption{Temporal variation of MPIW for baseline and STZITD-GNN models.}
\label{fig:mpiw_long}
\end{figure*}

Evaluating temporal stability is crucial for evaluating a model's performance over multiple prediction time steps, particularly in data-driven methods, where superior results in specific instances might mask potential underfitting issues \citep{ding2022deep, mannering2018temporal}. Our primary goal is to achieve consistently low uncertainty, as indicated by the MPIW metric in three boroughs, depicted in Figure \ref{fig:mpiw_long}. We strive to maintain a narrow yet comprehensive MPIW throughout the prediction steps to ensure reliable prediction intervals for traffic crashes.

Analysis shows that the MPIW of the top performing models, including STZITD-GNNs, is significantly lower than that of the baseline models, emphasising their superior performance throughout the study period. Therefore, we focus on the temporal dynamics of these leading models, presenting baseline performances as averages for contextual clarity.

As indicated in Figure \ref{fig:mpiw_long}, the MPIW is typically lowest on the first day of testing and generally increases over time. However, this increase is smoother in our STZITD-GNNs model compared to the more erratic fluctuations observed in baseline models. This smoother increase reflects our model's ability to handle the inherent uncertainties and the variability of risk levels, which might be less accurately captured by models with misspecified distributions. Moreover, boroughs like Westminster and Lambeth, known for higher instances of zero-crash risk values, show more fluctuating MPIWs, suggesting greater uncertainties in prediction intervals due to the sparsity of traffic crashes. Despite these challenges, our model consistently delivers stable and accurate predictions across different boroughs. 

\begin{figure}[!htp]
    \centering
    \begin{subfigure}[b]{1\textwidth}
        \includegraphics[width=\linewidth]{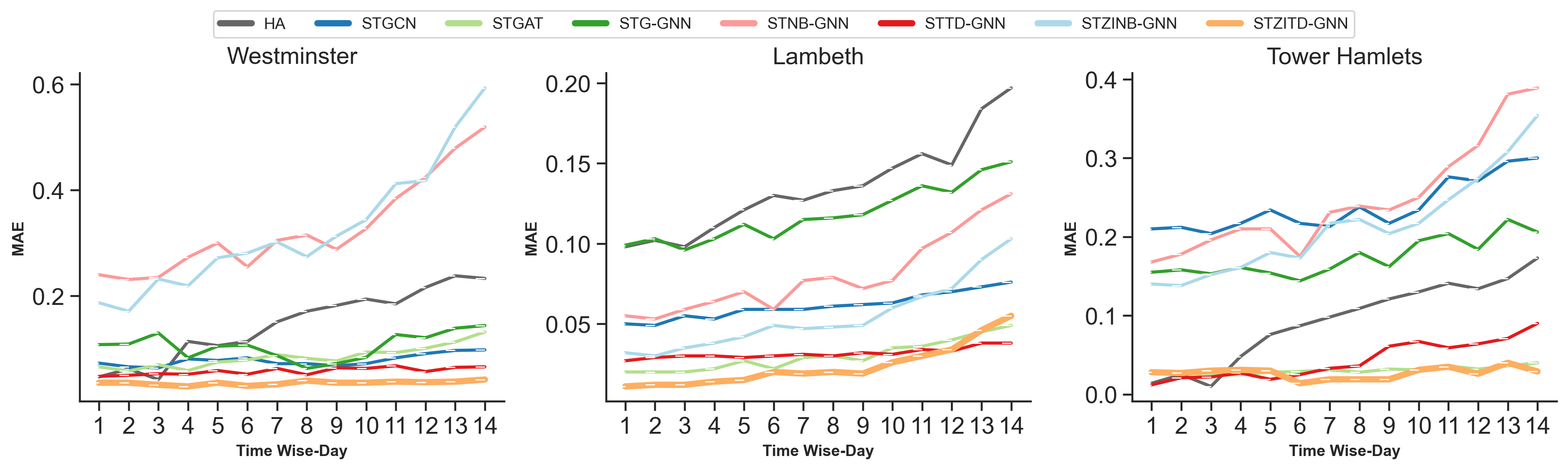}
        \caption{MAE over Time Performance}
        \label{fig:MAE}
    \end{subfigure}
    \hfill 
    
    \begin{subfigure}[b]{1\textwidth}
        \includegraphics[width=\linewidth]{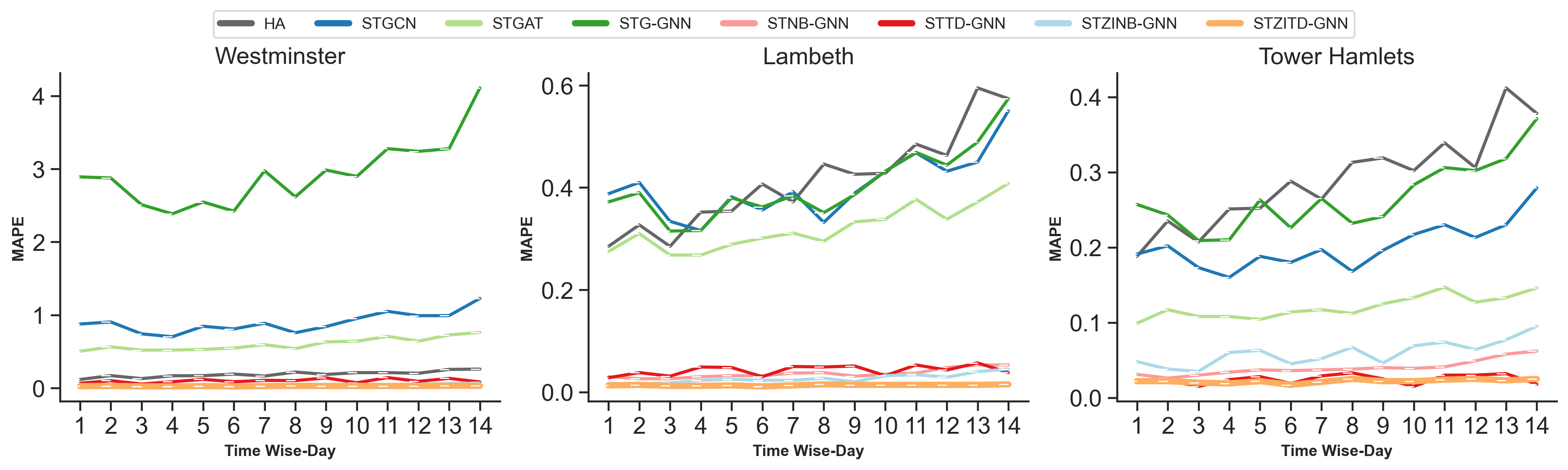}
        \caption{MAPE over Time Performance}
        \label{fig:MAPE}
    \end{subfigure}
    \hfill 

    \begin{subfigure}[b]{1\textwidth}
        \includegraphics[width=\linewidth]{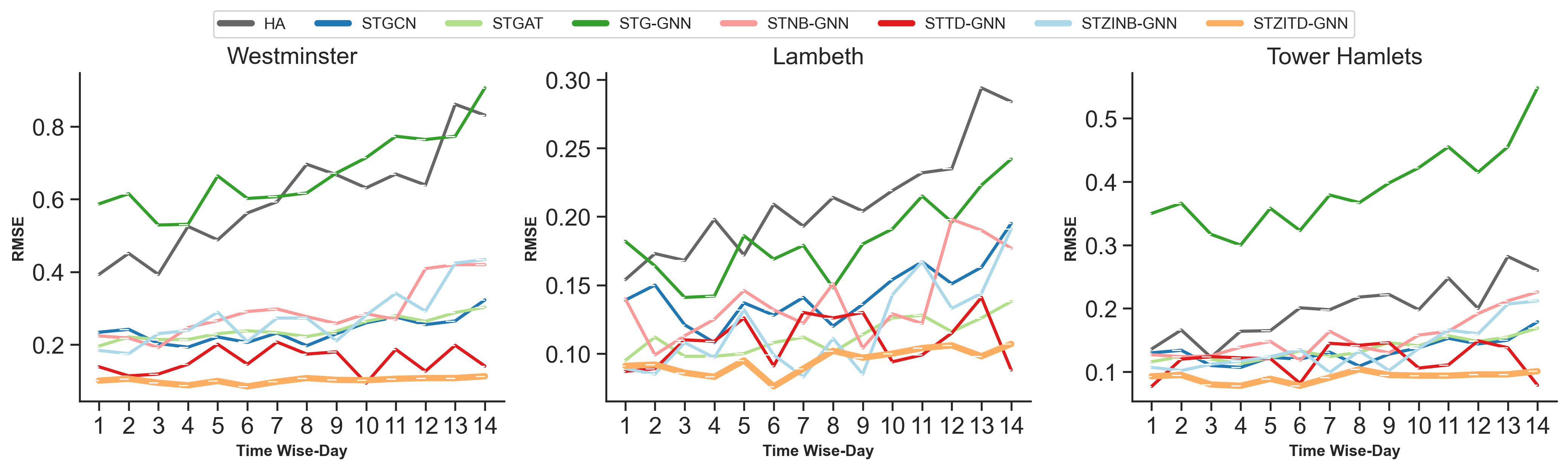}
        \caption{RMSE over Time Performance}
        \label{fig:RMSE}  
    \end{subfigure}
    \caption{Time-wise Comparison for the Point Estimation Metrics}
    \label{fig:time_class_point_metrics}
\end{figure}

In the evaluation of the temporal stability for the estimation of points, depicted in Figure \ref{fig:time_class_point_metrics}, the STZITD-GNN customised approach emphasises the advantages of applying the most appropriate probabilistic assumptions, enhancing spatiotemporal representations, and managing the various aspects of traffic crashes. It is apparent that the STG-GNN model consistently shows the weakest performance across all temporal prediction periods, with deterministic models also performing poorly. Furthermore, models based on the Negative Binomial (NB) distribution exhibit unsatisfactory performance characterised by higher fluctuation and higher error rates. This inadequacy stems from the focus of NB models on binary crash-occurrence outcomes, which often overlook the complexities of long-tail crash risk values. In contrast, our model exhibits significantly less variance across all three evaluated metrics at each time step, demonstrating stability even toward the end of the prediction horizon. This robust performance is attributed to the model's ability to dynamically fine-tune prediction intervals using four critical parameters, effectively addressing the challenges posed by the sparsity and skewed distribution of traffic crash data.

In terms of the zero rate, deterministic deep learning models tend to conflate no-risk with lower-risk values. This occurs because deterministic models do not adequately differentiate between the absence of risk and the presence of minimal risk. As a result, these models often misclassify low-risk scenarios as zero-risk, particularly in situations with extreme zero inflation, leading to overfitting. For the hit rate, the issue is inversely presented as low-risk predictions often being incorrectly classified as no-risk. This misclassification is due to the model's inability to capture subtle variations in risk levels.

\begin{figure}[!htp]
    \centering
    \begin{subfigure}[b]{1\textwidth}
        \includegraphics[width=\linewidth]{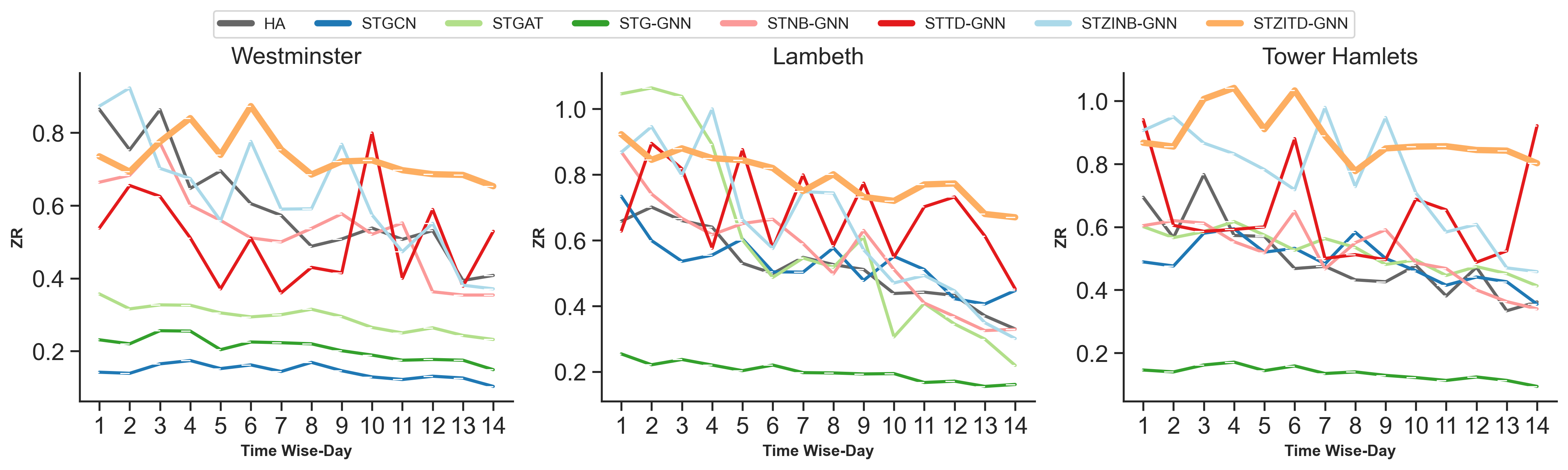}
        \caption{Zero Rate over Time Performance}
        \label{fig:ZR}
    \end{subfigure}
    \hfill
    
    \begin{subfigure}[b]{1\textwidth}
        \includegraphics[width=\linewidth]{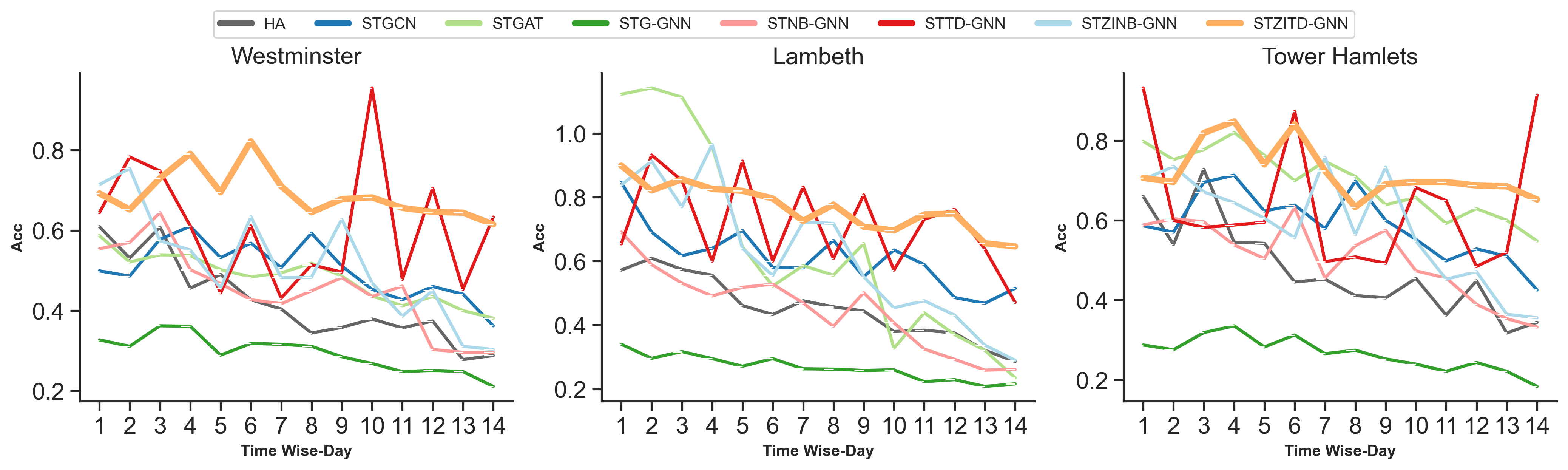}
        \caption{Accuate Hit Rate on Crash Occurrence at Top 20\% Risky Roads over Time Performance}
        \label{fig:Acc_at}
    \end{subfigure}
    \hfill
    \caption{Time-wise Comparison for the Non-crash Roads Quantification and Accurate Crash Hit Metrics}
    \label{fig:time_ZR_metrics}
\end{figure}

The final component of our evaluation focusses on quantifying both non-crash roads and high-risk roads where actual crashes occurred, as shown in Figure \ref{fig:time_ZR_metrics}. The STZITD-GNN model excels consistently in both metrics, maintaining stable performance with less than 18\% variance throughout the 14-step period. Unlike deterministic models, which exhibit less variability but struggle significantly with accurately identifying true noncrash roads, often misclassifying all roads as having no crash risk, the STZITD-GNN model effectively addresses these challenges.

In scenarios of extreme zero inflation, deterministic models frequently misclassify roads, conflating no-risk with lower risk values, thereby leading to overfitting. Conversely, probabilistic models using Gaussian distributions struggle to manage the high prevalence of zero values and lower risk values, resulting in poor performance and high sensitivity in the ZR metrics over time. This problem occurs because each time step may present unique scenarios of zero-valued roads, which Gaussian models often misinterpret due to their uniform mean and variance assumptions.

In the identification of actual crashes on high-risk crash roads, as illustrated in Figure \ref{fig:Acc_at}, the ZITD distribution performs exceptionally well. It employs a weighted dispersion parameter ($\phi$) that effectively distinguishes lower crash risks from non-crash roads, accurately reflecting the distribution of low, medium, and high-risk roads where actual crashes occurred, shown as long-tailed data distributions. This adaptability to the complex temporal dynamics of road-level crash data at each time step not only mitigates overfitting for non-crash scenarios, but also stabilises the model performance, enabling it to reliably capture real crash occurrences in critical geographic locations.

\subsubsection{Analysis of Long-tail Crash Data Characteristics}
\begin{figure}[!htbp]
\vspace{-0.5mm}
    \begin{minipage}{0.31\columnwidth}
		\centering
		\includegraphics[width=0.99\columnwidth]{ 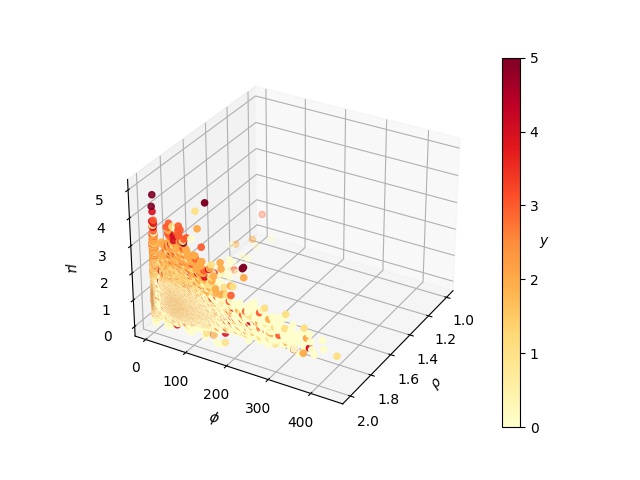}
		\subcaption{Westminster}
    \end{minipage}
    \begin{minipage}{0.31\columnwidth}
	\centering
        \includegraphics[width=0.99\columnwidth]{ 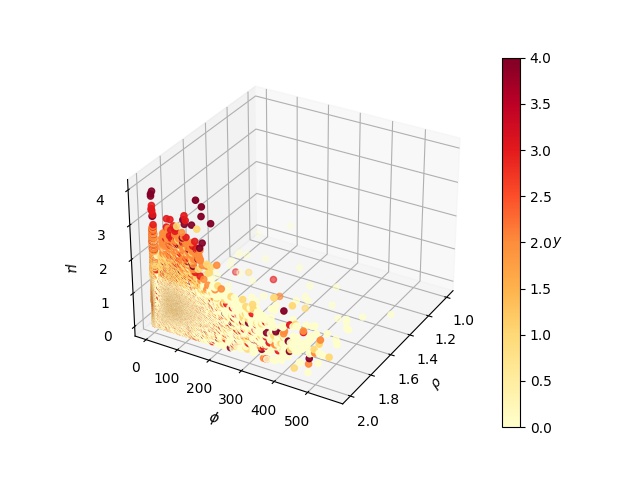}
	\subcaption{Lambeth}
	\end{minipage}
    \begin{minipage}{0.31\columnwidth}
		\centering
        \includegraphics[width=0.99\columnwidth]{ 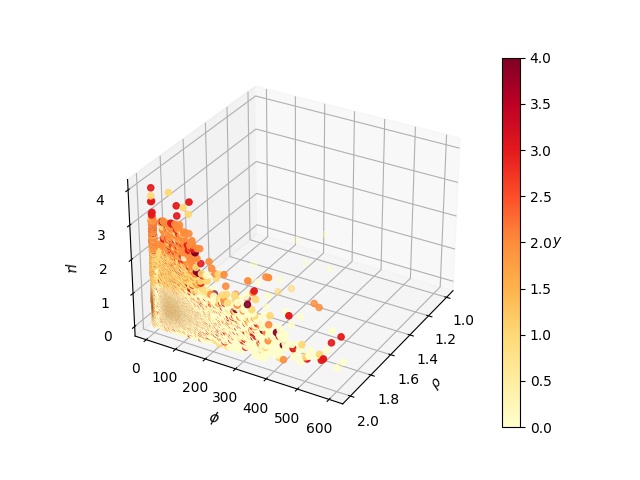}
			\subcaption{Tower Hamlets}
    \end{minipage}

	  \caption{Three Tweedie parameter visualization on Lambeth, Tower Hamlet, and Westminster with multi-step predict. The colour bar shows the predicted crash from the highest to zero.}
\label{fig:3d}
\end{figure}

To explore the impact of inherently long-tailed data on our model, we created a three-dimensional surface plot over four information sources, as illustrated in Figure \ref{fig:3d}. Due to visualisation constraints, it is not feasible to represent all four parameters in a single plot. We therefore focus on the TD parameters $\phi$, $\rho$, and $\mu$, with the understanding that the sparsity parameter $\pi$ significantly influences the original TD parameters. This plot illustrates the predicted crash values, with the actual crash values $y$ represented by the colour gradient.

A distinct trend is evident in the learnt values of $\rho$, consistently approaching 2, suggesting a preference for more skewed distributions as $\rho$ increases from 1 to 2 \citep{tweedie1984index, kurz2017tweedie}. This trend corresponds with the long-tailed nature of our data and validates the probabilistic assumptions. The increase in $\rho$ serves as an adaptive response to the component of the loss function $y\frac{\mu^{1-\rho}}{1-\rho} (y>0)$, effectively penalising the underestimation of true, although infrequent, higher values of crash risk within the long tail. This adjustment directs the model to focus more on such kinds of traffic crash rates, continually benefiting from significant enhancements in model sensitivity and accuracy.

Additionally, significant variations in the values $\phi$ are observed, particularly concerning data with zero value, which serve as the dispersion parameter. For example, in Westminster, $\phi$ is significantly lower than in other boroughs, indicating fewer zero-inflation issues. Traditionally, $\phi$ in the TD distribution is crucial for managing the frequency of zero occurrences to effectively address zero inflation. However, due to the highly imbalanced nature of our data, a single $\phi$ parameter often fails to capture the prevalence of zeros, as demonstrated by the overlap of plots in areas of lower and zero crash values. In our ZITD model, an enhanced $\phi$, augmented by the sparsity parameter $\pi$, manages these features, offering a precise depiction of the inherent sparsity and discrete uncertainty in road-level crash predictions.

Moreover, the model parameter $\mu$, which denotes the mean of the predicted crash values, shows a strong correlation with the actual crash values $y$, as evidenced by the colour alignment in the graphs. This correlation not only substantiates the effectiveness of the model in estimating mean points, but also affirms its capability to accurately mirror and adapt to the complexities of urban crash data.

\section{Conclusion and Discussion}
\label{sec:conclusions}
This paper introduces STZITD-GNN, a probabilistic deep learning framework that, to the best of our knowledge, is the first uncertainty-aware model specifically tailored for road-level crash prediction. Developed from data-driven insights, the STZITD-GNN utilises predefined distribution characteristics of traffic crash data to enhance the representation of sparse traffic crash scenarios. It effectively detects finer-level correlations in traffic crashes, thereby delivering precise predictions with narrow prediction intervals and reduced uncertainties. Through systematic multi-timescale experiments conducted on three real-life urban crash datasets, our results demonstrate that the efficacy of our STZITD-GNN approach surpasses that of state-of-the-art probabilistic deep learning models as well as deterministic models. The key contributions of this study are summarised as follows.

\begin{itemize}
    \item \textbf{Innovative Distribution Extension:} We have adapted the conventional TD to a ZITD distribution, introducing a sparsity parameter, $\pi$, integrated with the dispersion parameter $\phi$. This modification tailors the model to our data-driven assumptions, representing the dispersed and extreme zero-inflated nature of crash data at the road level. 
    
    \item \textbf{Integration with ST-GNN:} The incorporation of the ZITD encoder into the ST-GNN framework makes the STZITD-GNN model an end-to-end framework to handle the complex spatiotemporal dynamics of crashes. It surpasses traditional point estimation methods by enabling the mapping of output variance and prediction intervals. Additionally, the model facilitates an interpretable, uncertainty-aware process that elucidates the underlying factors contributing to improved prediction accuracy.
    
    \item \textbf{Empirical Validation Across Boroughs:} The model has shown improved predictive accuracy and minimal uncertainties compared with benchmark models in three  London boroughs, each featuring unique spatial crash patterns in multi-step scenarios. This performance confirms the robustness and generalizability of the model across diverse urban settings, even with sparse datasets. Additionally, the model's stable performance across heterogeneous datasets, which vary in crash determination criteria, further underscores its reliability and effectiveness in urban traffic risk assessment.

\end{itemize}

It should be recognised that the challenges of overdispersion and zero inflation extend beyond traffic crashes, affecting a wide array of urban issues such as crime rates. This universality suggests that our proposed model could be applicable to a broader range of contexts beyond its current usage.

While our research demonstrates its effectiveness in predicting traffic crashes, there are several limitations that provide direction for future research. We aim to enhance the accuracy of mapping crashes at finer temporal resolutions and plan to expand the model's applicability to various countries under consistent assumptions. Future developments will focus on creating a more adaptable model that can accommodate geographic differences, thus increasing its usability across diverse regions. In addition, we intend to integrate more diverse data types into our model. Incorporating GPS data, for example, could provide more detailed insight into traffic exposure. To further improve the accuracy and utility of our model, we also plan to use more comprehensive data sets, including urban characteristics derived from street or satellite images. This would enrich our model's contextual understanding and could identify factors that exacerbate traffic crash impacts across different locations and times. Addressing these challenges in future studies will improve the robustness and applicability of the framework in real-world scenarios, ultimately contributing to the development of more effective traffic management systems and crash prevention strategies.

In conclusion, the development of the STZITD-GNN model marks a significant advancement in the prediction of urban road-level traffic crashes. However, this is just the beginning of an ongoing effort to refine and enhance this model. By integrating additional data sources and continually improving our methodologies, our goal is to develop a tool that not only mitigates traffic risks but also significantly contributes to global urban safety strategies.

\section{Declaration of Generative AI and AI-assisted technologies in the writing process}
During the preparation of this work the author(s) utilized GPT-4 for preliminary grammar checks and proofreading. After using this tool/service, the author(s) reviewed and edited the content as needed and take(s) full responsibility for the content of the publication.

\clearpage
\bibliographystyle{apalike} 
\bibliography{ref}

\appendix
\section{Tweedie Table}
\label{app:td table}
\begin{table}[htbp!]
\centering
\begin{tabular}{c|ccccc}
\hline
\multicolumn{1}{c|}{Tweedie EDMs} & $\rho$                     & $\kappa(\theta)$                                              & $\phi$                      & $\theta$                       & $a(y, \phi)$                                           \\ \hline
Normal                             & \multicolumn{1}{c}{0}     & $\frac{\theta^2}{2}$                                          & $\sigma^2$                  & $\mu$                          & $\exp{(-y^2)/2\sigma ^2 - \log 2\pi / 2 )}$            \\
Poisson                            & \multicolumn{1}{c}{1}     & $\exp{(\theta)}$                                              & \multicolumn{1}{c}{$1$}    & $\log \mu$                     & $1/y!$                                                 \\

Poisson-Gamma                      & \multicolumn{1}{c}{(1,2)} & $\frac{(\theta-\rho \theta)^{\frac{2-\rho}{1-\rho}}}{2-\rho}$ & $\phi$                      & $\frac{\mu ^{1-\rho}}{1-\rho}$ & Eq.\ref{eq:td yy}                                                    \\
Gamma                              & \multicolumn{1}{c}{2}     & $-\log (-\theta)$                                             & \multicolumn{1}{c}{$\phi$} & $-1/\mu$                       & $\frac{\phi ^{-1/\phi} y ^{1/\phi-1}}{\Gamma(1/\phi)}$ \\ \hline
\end{tabular}
\end{table}
where $\mu, \sigma$ is the mean and the standard values of Gaussian Distribution.

\section{Tweedie Parameters}
\label{app:parameter}
The frequency of traffic crashes $C_{k} $ adheres to a Poisson distribution represented by $Pois(\lambda)$ with a mean of $\lambda$, expressed as $C_{k} \sim Pois(\lambda)$. This is natural, as frequency often appears as the main prediction target in the previous literature \citep{zhou2020riskoracle,trirat2023multi,bonera2024network}. Currently, the severity of the crash $l_{k}^{(j)}$ is modelled as independent and identically distributed $i.i.d$ Gamma random variables, $C_{k} \perp l_{k}^{(j)}$ and $\forall j$, denoted as $Gamma(\alpha, \gamma)$. These variables have a mean of $\alpha \gamma$ and a variance of $\alpha \gamma ^2$, such as $l_{k}^{(j)} \stackrel{iid}{\sim} \operatorname{Gamma}(\alpha, \gamma)$. $C_k$ and $l_{k}^{(j)}$ together form the model of the occurrence of a single crash. Consequently, the traffic crash risk $y_k$ is formulated as a Poisson sum of these i.i.d. Gamma random variables. Considering Equation \ref{eq: y_it} and considering $1<\rho<2$, we can finally reformulate the random variable $y_k$ as follows:

\begin{equation}
    y_k = \left\{ \begin{array}{cc}
        0 & \text{if } C_{k}=0,  \\
        \sum_{j=1}^{C_{k}} l_{k}^{(j)} = l_{k}^{(1)} + l_{k}^{(2)} + \cdots + l_{k}^{(C_{k})} & \text{if } C_{k}>0.
    \end{array}\right..
    \label{eq:td yy}
\end{equation}

In the scenario where $C_{k} =0$, we correspondingly have $y_k=0$. The probability density at zero for traffic crash risk, therefore, becomes $\mathbb{P}(y_{k}=0)=\mathbb{P}(C_{k}=0)=\exp (-\lambda)$. In contrast, when $C_{k}>0$, $y_k$ is deduced from the sum of $C_{k}$ independent Gamma random variables. For clarity, following the work of \citet{zhou2022tweedie}, we re-parameterize the TD distribution, setting $\theta = \mu^{1-\rho}/(1-\rho)$ and $\kappa(\theta)=\mu^{2-\rho}/(2-\rho)$. This results in the revised TD distribution as:
\begin{equation}
f_{\text{TD}}(y_k|\theta, \phi) =
f_{\text{TD}}(y_k|\mu, \phi, \rho)
= a(y_k, \phi, \rho)\exp {\bigl[ \frac{1}{\phi}({y_k \frac{\mu^{1-\rho}}{1-\rho}-\frac{\mu^{2-\rho}}{2-\rho}}) \bigl]},
\label{eq: re_f}
\end{equation}
where
\begin{equation}
    a(y_k, \phi, \rho) = \left\{ \begin{array}{cc}
        1 & \text{if } y_{k}=0,  \\
        \frac{1}{y_{k}} \sum_{j=1}^{\infty} 
        \frac{y_{k}^{-j \alpha}(\rho-1)^{\alpha j} }{\phi^{j(1-\alpha)} (2-\rho)^j j! \Gamma(-j\alpha)}
         & \text{if } y_{k}>0.
    \end{array}\right..
\end{equation}

In this case, $\mu, \phi$ and $\rho$ are the three parameters that govern the probability and expected value of the risk of traffic crashes. Parameters in the Gamma and Poisson distributions ($\lambda, \alpha, \gamma$) can be obtained as \citep{kurz2017tweedie, halder2019spatial,mallick2022differential}, :

\begin{equation}
\lambda = \frac{1}{\phi} \frac{\mu ^ {2-\rho}}{2-\rho}, \quad \alpha = \frac{{2-\rho}}{\rho-1}, \quad \gamma=\phi (\rho-1) \mu ^ {\rho-1}.
\label{eq: lambda, alpha, gamma}
\end{equation}

\section{Temporal encoder -- GRU}
\label{app:GRU}
Traffic crashes demonstrate a significant temporal correlation, underscoring the idea that the dynamic temporal features of the roads encapsulate a wealth of information. This observation has been acknowledged and exploited in prior methodologies \citep{wang2021gsnet, zhao2019t}. As shown in Figure \ref{fig:method}, to seize these temporal proximity features, we implement a GRU \citep{GRU}. The GRU, in comparison to traditional LSTM networks, offers the advantage of learning relatively long-term dependencies without facing issues of vanishing and exploding gradients. In the GRU framework, there isn't a segregation of internal and external states, as seen in LSTM networks. Instead, it addresses the gradient vanishing and exploding problem by directly integrating a linear dependency between the current network state $h_t$ and the preceding state $h_{t-1}$. Consequently, while a GRU preserves the capabilities of an LSTM network, its architecture remains more streamlined (\cite{gu2019short,mahmoud2021predicting,ma2022novel}).

Specifically, this component accommodates the feature sequence over past time windows $1:t$, thereby learning temporal dependencies. We represent the temporal embedding as $\mathcal{Z}_T$, taking into account both historical spatiotemporal road features and traffic crashes as input:
\begin{equation}
\centering
 \begin{aligned}
\mathcal{Z}_T = \text{GRU}(X_{1:t},Y_{1:t}),
    \label{eq:GRU Total}
 \end{aligned}
\end{equation}

Once the input features have been processed via the GRU, we obtain historical temporal embeddings $\mathcal{Z}T$ for all roads. These are subsequently directed to graph neural encoders. The GRU, in particular, consists of two gates: the update gate and the reset gate. The encoder processes a single input $X_{t}, Y_{t}$ at time slot $t$ and prior hidden features $h_{t-1}$ in sequence:
\begin{equation}
\centering
 \begin{aligned}
& r_t = \sigma (W_r \cdot [h_{t-1}, X_{t}, Y_{t}] + b_r),
\\
& u_t = \sigma (W_u\cdot [h_{t-1}, X_{t}, Y_{t}] + b_u),
\\
& \tilde{h}_t = \text{tanh}(W_c \cdot [r_t \cdot h_{t-1}, X_{t}, Y_{t}] + b_c),
\\
& h_t = (1-u_t)\cdot h_{t-1} + u_t \cdot \tilde{h}_t
    \label{eq:GRU}
 \end{aligned}
\end{equation}
In the equation above, $h_{t}$ and $h_{t-1}$ represent hidden features at the current and preceding time steps, respectively, while $\tilde{h}_t$ denotes the candidate state that temporarily stores information from reset and update gates ($r_t$,$u_t$). The variables $W$ and $b$ (i.e. $W_r, W_u, W_c, b_r$, $b_u, b_c$) represent learnable weight and bias matrices, respectively. Importantly, $r_t$ acts as the reset gate, determining the extent of integration of hidden features from previous time steps into features at the current time step. Meanwhile, $u_t$ serves as the update gate, controlling how much information from previous time steps will be carried over to the next time step. The $\text{tanh}(\cdot)$ is the hyperbolic tangent function. We perform the GRU gate operations step by step and select the final layer's hidden features $h_t$ as the temporal embedding $\mathcal{Z}_T \in \mathbb{R}^{N\times F}$, where $F$ denotes the dimension of road temporal embedding.

\section{Spatial encoder -- GAT}
\label{app:GAT}
Within our research, road data is fundamentally perceived as having an inherent graph structure. This perspective facilitates the interpretation of node features as graph signals. To accommodate such graph-structured information, we employ the GAT \citep{GAT}, an effective methodology renowned for enabling potential model deepening while concurrently incorporating dynamic spatial correlation into our network, achieved through real-time adjustment of attention weights in response to data changes \citep{wang2022attention,zhang2019spatial}.

Distinct from GCN of a spectral-domain convolution \citep{kipf2016semi}, GAT operates as a spatial-domain convolutional network. Instead, GAT emphasizes the magnitude of the graph signal value at distinct points and their inter-distance to execute the convolution process. This trait empowers GAT to alleviate the over-smoothing issue associated with GCN deepening, thereby augmenting prediction precision regarding peaks and valleys \citep{zhao2019t}.

Initially, we compute the attention coefficient $\alpha_{ij}$ between roads $v_i$ and $v_j$ as per the following equation:

\begin{equation}
    \centering
    \begin{aligned}
    \alpha_{ij} = \frac{\exp({\text{LeakyReLU}\bigl(\vec{a}^\mathsf{T}[W_a\mathcal{Z}{T_i}||W_a{\mathcal{Z}{T_j}])\bigl)}}}{\sum_{A_{i,o}>0}\exp({\text{LeakyReLU}\bigl(\vec{a}^\mathsf{T}[W_a{\mathcal{Z}{T_i}}||W_a{\mathcal{Z}{T_o}}])\bigl)}},
    \label{eq:attention}
    \end{aligned}
\end{equation}

In this equation, $A_{i,o}>0$ represents the neighbourhood or connected roads of the road $v_i$. $W_a\in \mathbb{R}^{F \times F'}$ and $\vec{a} \in \mathbb{R}^{2F'}$ are learnable weights, where $F'$ denotes the hidden dimension. We apply the concatenation function $[\cdot||\cdot]$ to concatenate the embedding of $v_i$ and its neighbours, utilizing learnable parameter $W_a$ and mapping function $\vec{a}$. Subsequently, we employ an activation function, $\text{LeakyReLU}(\cdot)$, and normalize the attention coefficient $\alpha_{ij}$ using the $\exp (\cdot)$ function. The embedding of road $v_i$ is consequently updated with a linear combination of the neighbourhood of road $v_i$ and itself, wherein these computed attention coefficients are employed.

In order to preemptively address potential inaccuracies in attention assignments, we have incorporated multiple attention assignments, otherwise known as multiple attention heads, into our approach. This tactic greatly amplifies the expressive capacity of the model. The multi-head attention mechanism facilitates concurrent consideration of information from a wide range of representation subspaces at various positions. This enables each head to learn and focus on different facets of the input data, thereby enhancing the model's ability to comprehend and accurately represent the data \citep{reza2022multi,tang2022spatiotemporal,ding2023mst}. Herein, $M$ symbolizes the number of attention mechanisms, with the output embeddings from these mechanisms being concatenated in the following manner:

\begin{equation}
    \centering
    \begin{aligned}
    \mathcal{Z}_{i} = ||_{m=1}^M \sigma\left(\sum_{A_{i,j}>0} \alpha_{ij}^m {W_a}^m \mathcal{Z}_{T_j}\right),
    \label{eq:multi-head}
    \end{aligned}
\end{equation}

In this equation, superscript $m$ denotes a different head. The embedding of the road $v_i$ is optimized by aggregating its neighbours' embedding with attention coefficients $\alpha_{ij}^{m}$, and we apply learnable weights ${W_a}^m$ for different heads. After this, we apply a nonlinear sigmoid function and concatenate different heads of embedding. The multi-head attention mechanisms hereby ensure stability in the attention mechanisms learning process and encapsulate diverse perspectives of road-level graph features. Notably, in Eq \ref{eq:multi-head}, the dimension of the output feature is $M\times F'$, rather than $F'$. To ensure consistency in the output feature's dimension, we compute the average of each road's embedding in the last layer as follows:

\begin{equation}
    \centering
    \begin{aligned}
    \mathcal{Z}_{i} = \sigma\left(\frac{1}{M}\sum_{m=1}^M \sum_{A_{i,j}>0} \alpha_{ij}^m {W_a}^m \mathcal{Z}_{T_j}\right).
    \label{eq:multi-head last layer}
    \end{aligned}
\end{equation}

Finally, we input temporal embedding $\mathcal{Z}_{T}$ into GAT to capture spatial dependency and obtain spatiotemporal $\mathcal{Z}_{i} \in \mathbb{R}^{F'}$ of road $v_i$. After that, we use $\mathcal{Z}$ to infer the four-parameter STZITD-GNNs model.

\end{document}